\documentclass{sig-alternate}

\usepackage[
  pass,
]{geometry}

\usepackage{algorithm}
\usepackage{algpseudocode}
\usepackage{amsmath}
\usepackage{graphics}
\usepackage{epsfig}
\usepackage{subfigure}
\usepackage{graphicx,dblfloatfix}
\usepackage{float}
\usepackage{multirow}
\usepackage{tabularx}
\usepackage{amsfonts,amssymb}
\usepackage[table,xcdraw]{xcolor}
\usepackage[pagebackref=true,breaklinks=true,letterpaper=true,colorlinks,citecolor=blue,linkcolor=blue,bookmarks=false]{hyperref}
\usepackage{xspace}

\makeatletter
\def\hlinewd#1{%
  \noalign{\ifnum0=`}\fi\hrule \@height #1 \futurelet
   \reserved@a\@xhline}
\makeatother

\makeatletter
\DeclareRobustCommand\onedot{\futurelet\@let@token\@onedot}
\def\@onedot{\ifx\@let@token.\else.\null\fi\xspace}

\def\ie{\emph{i.e}\onedot}

\def\etal{\emph{et al}\onedot}
\makeatother

\begin{document}
%
\conferenceinfo{ACM Multimedia}{'15 Brisbane, Australia}

\title{To Know Where We Are: Vision-Based Positioning in Outdoor Environments}
%
%
%
%
%

\numberofauthors{5} 
%
\author{
%
%
\alignauthor
Kuan-Wen Chen\\
       \affaddr{Intel-NTU Connected Context Computing Center}\\
       \affaddr{National Taiwan University, Taiwan}\\
       \email{kuanwenchen@ntu.edu.tw}
\alignauthor
Chun-Hsin Wang, Xiao Wei, Qiao Liang\\
       \affaddr{Graduate Institute of Networking and Multimedia}\\
       \affaddr{National Taiwan University, Taiwan}\\
\and
\alignauthor 
Ming-Hsuan Yang\\
       \affaddr{Electrical Engineering and Computer Science}\\
       \affaddr{University of California at Merced, USA}\\
       \email{mhyang@ucmerced.edu}
\alignauthor 
Chu-Song Chen\\
       \affaddr{Institute of Information Science}\\
       \affaddr{Academia Sinica, Taiwan}\\
       \email{song@iis.sinica.edu.tw}
\alignauthor 
Yi-Ping Hung\\
       \affaddr{Graduate Institute of Networking and Multimedia}\\
       \affaddr{National Taiwan University, Taiwan}\\
       \email{hung@csie.ntu.edu.tw}
}
\additionalauthors{Additional authors: John Smith (The Th{\o}rv{\"a}ld Group,
email: {\texttt{jsmith@affiliation.org}}) and Julius P.~Kumquat
(The Kumquat Consortium, email: {\texttt{jpkumquat@consortium.net}}).}
\date{30 July 1999}

\maketitle
\begin{abstract}
Augmented reality (AR) displays become more and more popular recently, because of its high intuitiveness for humans and high-quality head-mounted display have rapidly developed. To achieve such displays with augmented information, highly accurate 
image registration or ego-positioning are required, but little attention have been paid for outdoor environments.
This paper presents a method for ego-positioning in outdoor environments with low cost
monocular cameras. 
To reduce the computational and memory requirements as well as the
communication overheads, we formulate the model compression algorithm
as a weighted k-cover problem for better preserving model
structures. 
Specifically for real-world vision-based positioning applications, we
consider the issues with large scene change and propose a model update
algorithm to tackle these problems.
A long-term positioning dataset with more than one month, 106 sessions,
and 14,275 images is constructed.
Based on both local and up-to-date models constructed in our approach,
extensive experimental results show that high positioning accuracy ($mean \sim 30.9cm,  stdev. \sim 15.4cm$) can be achieved, which outperforms existing vision-based algorithms. 
\end{abstract}




\section{Introduction}
\label{sec:introduction}

\begin{figure*}[t]
\begin{center}
\subfigure[]{
\includegraphics[width=0.45\linewidth]{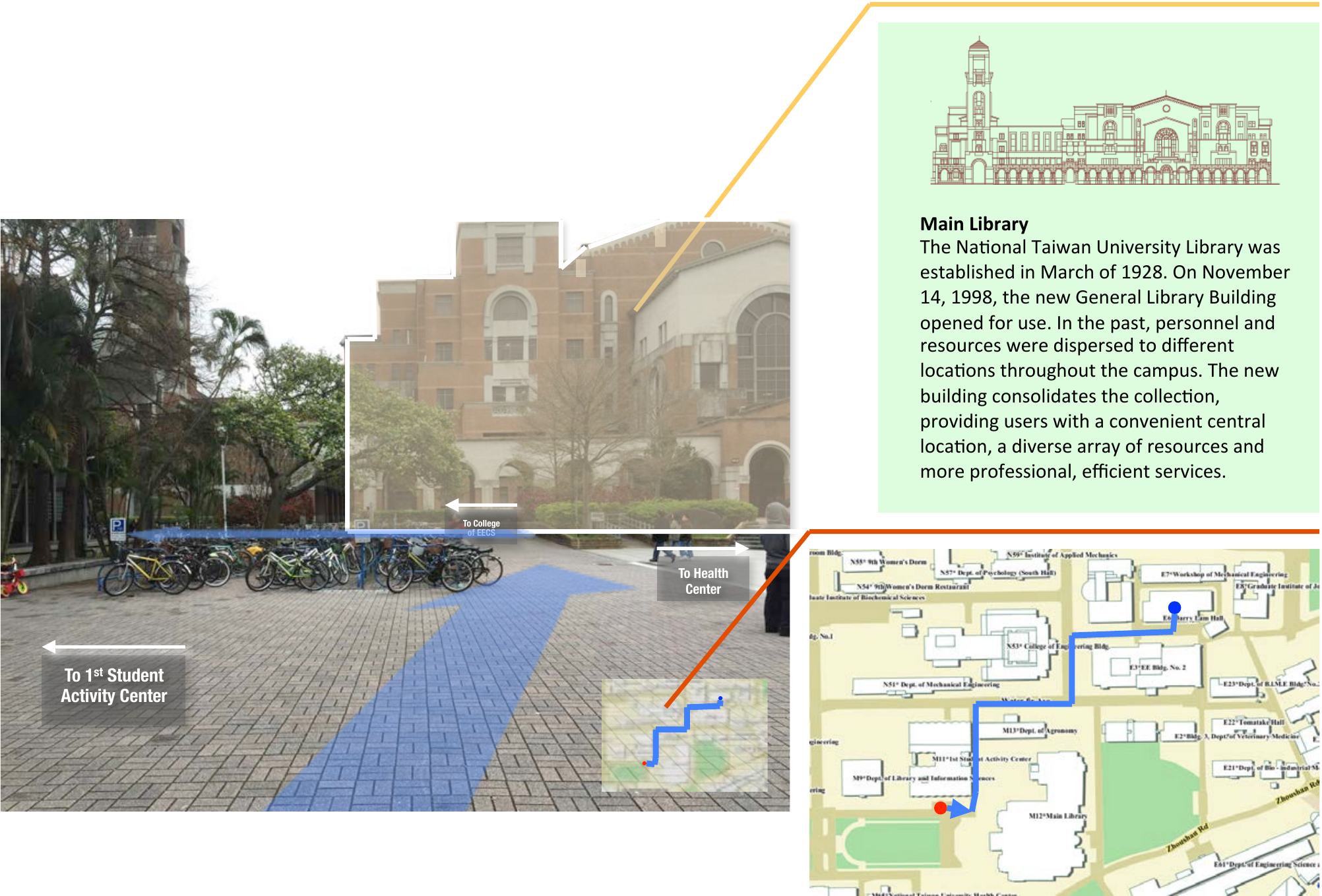}}
\subfigure[]{
\includegraphics[width=0.45\linewidth]{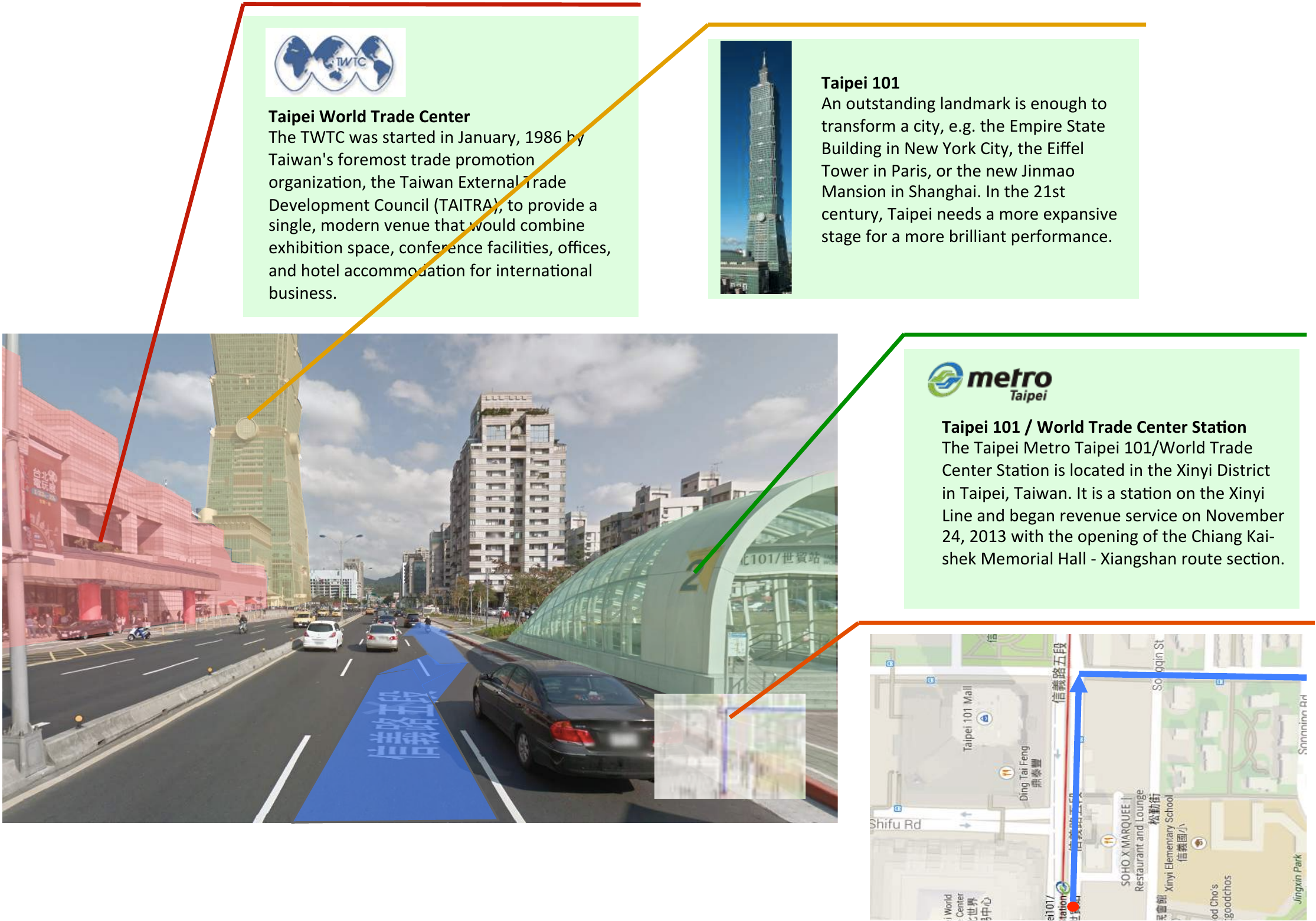}}
\end{center}
   \caption{Examples of applications while highly accurate positioning is available: (a) a natural way to provide buildings, shopping, or navigation information with AR technologies on mobile phone or head-mounted display, and (b) to provide fine-grained navigation, such as changing lane, on head-up display of vehicles.}
\label{fig:App}
\end{figure*}

Ego-positioning aims at locating an object in a global coordinate
system based on the sensors mounted on the object. 
With the growth of mobile or wearable devices, highly accurate positioning becomes
increasingly important. 
For example, Microsoft HoloLens \cite{MSHoloLens} uses AR technologies to display information in a natural way and enable users to interact with three-dimensional holograms. Pioneer NavGate HUD \cite{NavGateHUD} augments traffic and navigation information on a head-up display.
Figure~\ref{fig:App} shows another two examples of AR applications on mobile devices while highly accurate positioning is available. It enables display information on the corresponding area of image and helps users to connect virtual information with real world. An intuitive and fine-grained navigation suggestion can be provided instead of macro suggestion, such as turn right at the next intersection, which provided by the traditional navigation system but sometimes difficult to be understood when the road is complicated. To achieve such AR displays, highly accurate image registration or ego-positioning will be necessary.

Unlike indoor positioning \cite{GoogleTango,Gu,Koyuncu,Liu07,Martin10},
less effort has been made on high-accurate ego-positioning for outdoor
environments, and GPS (Global Positioning System) is still the most popular
positioning technology widely used. 
However, the precision of GPS sensors is around 3 to
20 meters \cite{Driver,Modsching} and existing systems do not perform
well in urban areas full of high rises. 
Although several positioning methods based on expensive sensors such
as radar and Velodyne 3D laser scanners can achieve high accuracy,
it is currently not practical to deploy them in mass, especially for low-cost mobile devices.

\begin{figure}[t]
\begin{center}
\includegraphics[width=\linewidth]{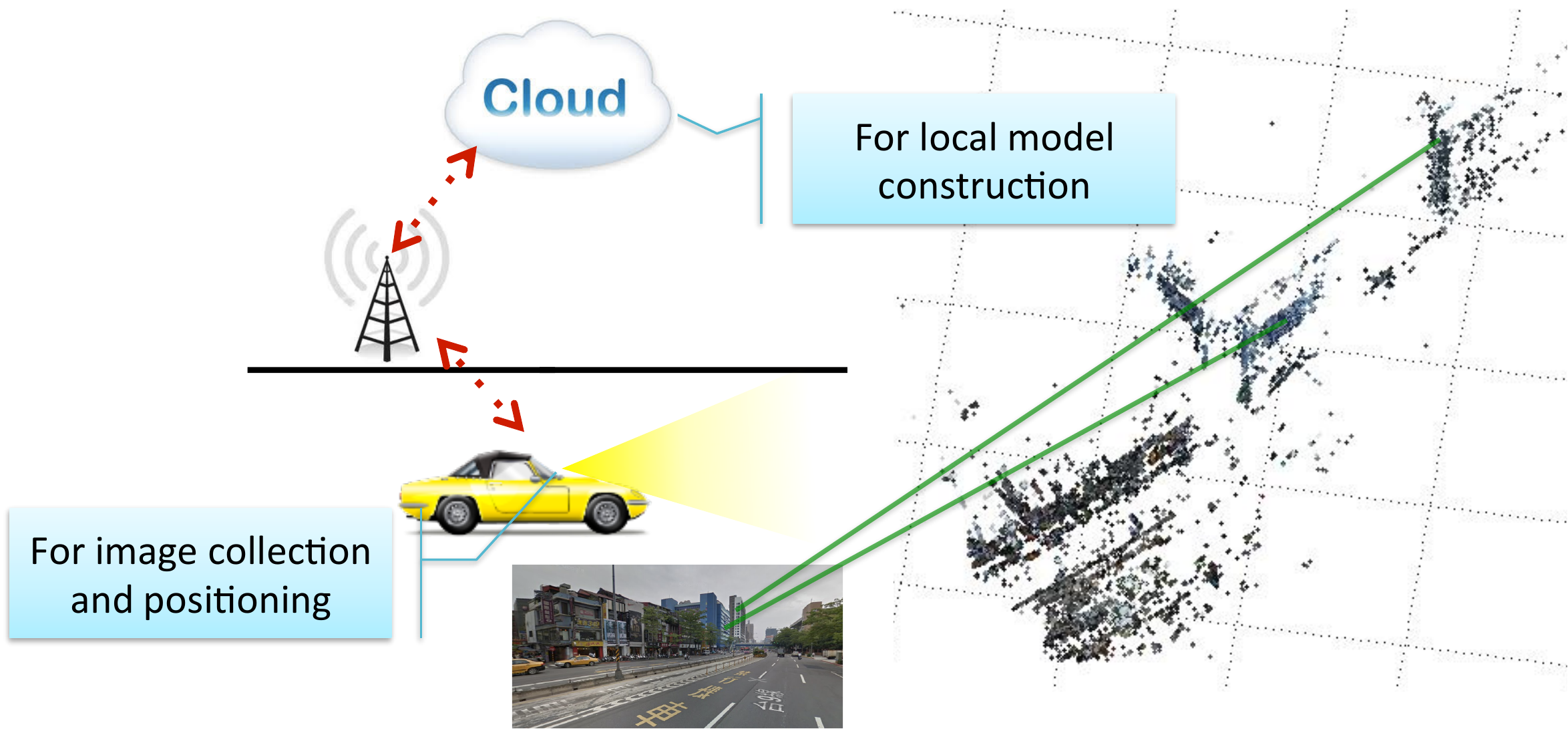}
\end{center}
   \caption{Overview of the proposed vision-based positioning
system.
SIFT features from images acquired on vehicles are matched against 3D 
models previously constructed for ego-positioning. 
In addition, the newly acquired images are used to update 3D models.}
\label{fig:Overview}
\end{figure}

In this paper, we introduce a vision-based positioning algorithm based
on low-cost monocular cameras within the IoT (Internet-of-Things) framework. 
It exploits visual information from the
images captured by mobile devices to achieve high positioning accuracy
even in crowded traffic situations. 
An overview of the proposed algorithm is shown in
Figure~\ref{fig:Overview}.  

To this end, our approach includes the training and 
ego-positioning phases. 
In the training phase, if a person/vehicle passes a
specific area (e.g., an intersection) that can be roughly positioned by
GPS, the captured images will be uploaded to a cloud system. 
After many persons/vehicles passing by that
area, the cloud system will collect a sufficient number of images to
construct the local 3D point set model of the area by using
a structure-from-motion algorithm~\cite{Snavely}. 
In the ego-positioning phase, the device on human/vehicle will start
to download all the local 3D scene models along the route. 
When en route, the GPS system informs the person/vehicle its rough position,
and the mobile device will match its current image with the
corresponding local 3D model for ego-positioning. 
Notice that our system only needs the
device to download the scene models in the ego-positioning phase, and
upload its own images for model construction and update when the wireless bandwidth is available.

Model-based positioning~\cite{Li12,Liu12,Sattler11,Sattler12}
have been studied in recent years, but most approaches construct a single
city-scale or close-world model and focus on fast correspondence
search. 
Such a large-scale model makes the correspondence matching task 
difficult, and thus the rate of successfully registered images is
under 70\% and high positioning accuracy is difficult achieve 
(positioning error is with median of 1.6m and mean of 5.5m in~\cite{Li12}, and with median of 1.3m and mean of 15m in~\cite{Sattler11})
Furthermore, model-based positioning approaches always suffer from two
main problems: how to build a large-scale (or world-scale) model and
how to make the model up-to-date. 

The main contribution of this work are summarized as follows. 
First, we propose a novel algorithm within the IoT framework. 
By collecting new observations from previous passing person/vehicle, it solves the above-mentioned problems regarding model construction and makes the model-based positioning practical.
A local and up-to-date model is constructed that facilities 
sub-meter positioning accuracy.
Second, a novel model compression algorithm formulated as a weighted
k-cover problem to preserve model structure is developed. 
Third, a model update 
method is presented and a dataset including more than 14,000 images
from 106 sessions for more than one month (including sunny, cloudy,
rainy, night scenes) is constructed. 
To the best of our knowledge, the proposed system is first to consider
large outdoor scene changes over a long period of time for model-based positioning
with sub-meter accuracy. 

\section{Related Work}
\label{sec:related_work}
Vision-based positioning aims to match current images with stored 
images or pre-constructed models for relative or global pose
estimation. 
Such algorithms can categorized into three types  according to the
registered data for matching. i.e., consecutive frames, 
images in a database, and 3D models. 
In addition, we discuss related work on compression of 3D models and long-term positioning. 

\subsection{Vision-Based Positioning}

\subsubsection{Matching with Consecutive Frames}

Positioning methods by matching the current with previous
frames to estimate relative motion are widely used in
robotics \cite{Kneip,Lategahn,Schmid}, which is also known as visual
odometry. 
%
These methods combine the odometry sensor readings and visual data
from a monocular or stereo camera to estimate local motion
incrementally. 
The main drawbacks are only relative motion can be estimated and the
accumulated errors is large (about 1 meter for every 200 meters movement
\cite{Schmid}) with the drifting problems. 
In \cite{Brubaker} Brubaker \etal incorporate the road maps to 
alleviate the problem with accumulated errors.
However, the positioning accuracy is low (i.e., localize objects with
up to 3 meter accuracy). 

\subsubsection{Matching with Database Images} 
Methods of this category match images with those in a database to determine
the current position \cite{Chen,Zamir}. 
Such approaches are usually used in multimedia applications where
accuracy is not of prime importance, such as 
non-GPS-tagged photo localization and landmark identification. 
The positioning accuracy is usually low (between 10 and hundred of 
meters).

\subsubsection{Matching with 3D Models}
Methods in this category are based on a constructed 3D model. 
Arth \etal \cite{Arth} propose a method to
combine sparse 3D reconstructions and manually determine visibility
information to estimate camera poses in an indoor
environment. 
Wendel \etal \cite{Wendel} extend this method to a micro
aerial vehicle for localization for the outdoors scenarios. 
Both these methods are evaluated in limited spatial domains within a
short amount of time. 

For large-scale localization, one main challenge is how to match
features efficiently and effectively. 
Sattler \etal \cite{Sattler11}
propose a direct 2D-to-3D matching method based on visual vocabulary
quantization and a prioritized correspondence search to speed up the
feature matching process.
This method is extended to both 2D-to-3D and
3D-to-2D search for more efficiency \cite{Sattler12}. 
Lim \etal \cite{Lim} propose to extract more efficient descriptors for
every 3D point across multiple scales for scale invariance in feature
matching. 
Li \etal \cite{Li12} further deal with a worldwide scale
problem including hundreds of thousands of images and tens of millions of 3D
points. 
%
As these methods focus on an efficient correspondence search, 
the sheer large number of 3D point cloud model leaded to 
registration rate of 70\% and mean of localization error of 5.5
meter~\cite{Li12}.
These methods are thus not applicable to AR displays on mobile devices due to position
accuracy, computational and memory requirements.  

Recently, methods that use a combination of local visual odometry and
global model-based positioning are proposed \cite{Middelberg,Ventura}. 
%
These systems estimate relative poses of mobile devices and 
carry out global image-based localizations on remote servers to
overcome the problem of accumulative errors. 

However, all of these model-based positioning approaches do not consider outdoor scene changes, and evaluate the images
taken in the same session when the model is constructed. 
In this paper, we present a system that deals with large
scene changes, and update models for long-term positioning within the
IoT context. 

\subsection{Model Compression}
\label{sec:compression}

In addition to positioning, compression methods 
have been developed to reduce the computational and memory
requirements for large-scale models~\cite{Cao14,Irschara,Li10,Park}.
Irschara \etal \cite{Irschara} and Li \etal \cite{Li10} use a
greedy algorithm to solve a set cover problem to minimize the number
of 3D points while ensuring a high probability of successful
registration. 
Park \etal \cite{Park} formulate
the set cover problem as a mixed integer quadratic
programming problem to obtain an optimal 3D point subset. 
Cao and Snavely \cite{Cao14} further take into account both coverage and distinctiveness and propose a probabilistic approach to yield better performance. 
However, these approaches do not consider the structure information and suffer from the problem of uneven spatial distribution, which leads to worse pose estimation. 
To overcome this problem,
a weighted set cover problem is proposed in this paper to ensure 
the reduced 3D points fairly distributed in all planes and
lines, thereby facilitating accurate positioning.  

\subsection{Long-Term Positioning}
\label{sec:longterm}

Little attention has been paid for long-term positioning. Although there are some works \cite{Johns13,Johns13Feature} in robotics considering outdoor scene changes for localization and mapping, they only match 2D query images with dataset images collected from different period of time. Correct matching is considered the distance between both images is less than 10 or 20 meters by GPS measurement instead of estimating its 3D absolute positions as what we do in this paper.

\begin{figure}[t]
\begin{center}
\includegraphics[width=1\linewidth]{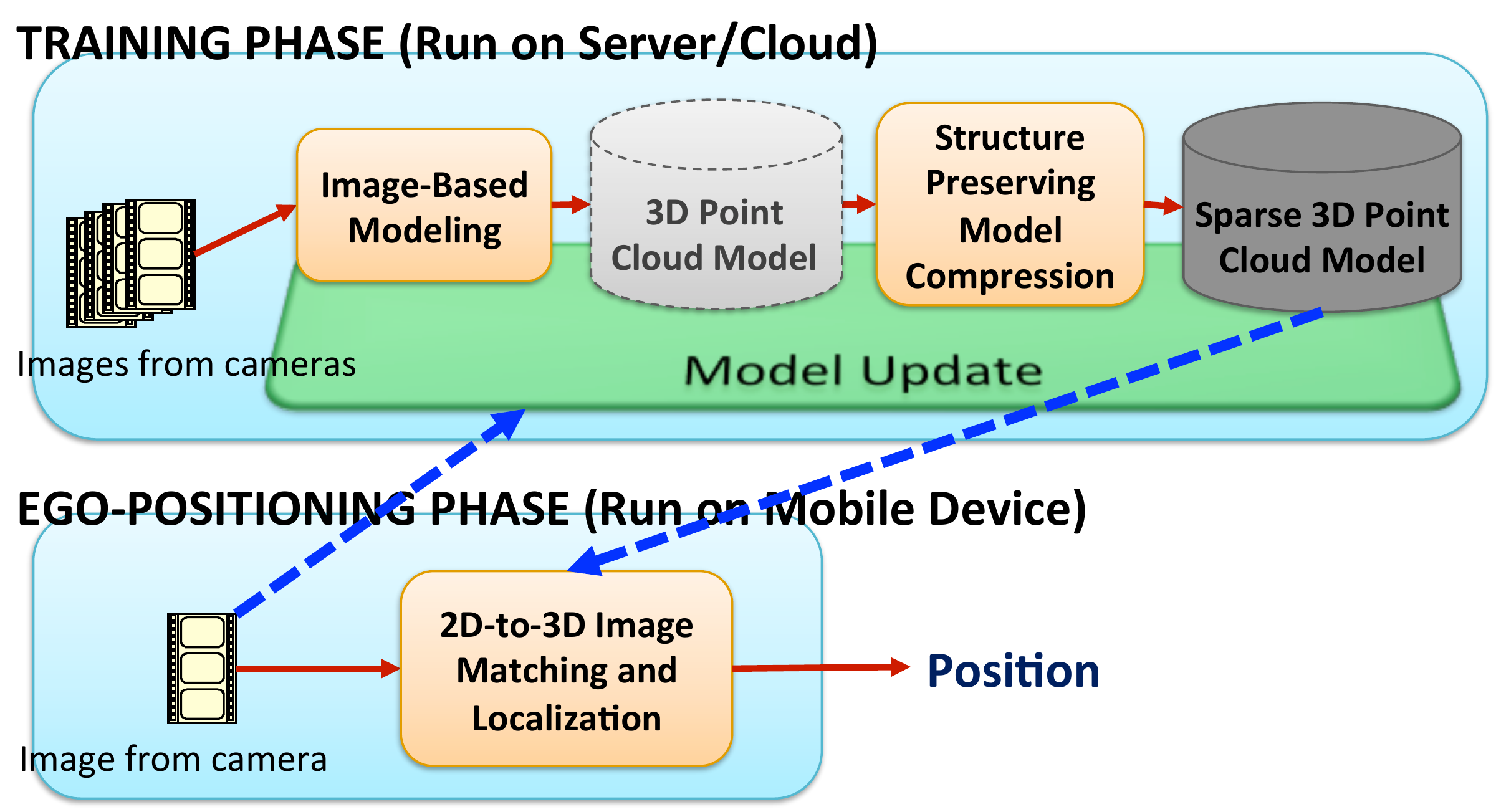}
\end{center}
   \caption{Overview of the proposed vision-based
     positioning system. 
The blue dot lines denote wireless communication links.}  
\label{fig:SystemOverview}
\end{figure}

\section{System Overview}
\label{sec:overview}

As discussed in Section~\ref{sec:introduction}, the proposed system
consists of two phases. 
Figure~\ref{fig:SystemOverview} shows the system overview. 
The training phase is carried out on local machines or cloud
systems. 
It includes three main components, and all of
them are processed off-line with batch processing. 
First, image-based modeling builds a 3D point cloud model
from collected images automatically. 
Second, structure preserving model
compression reduces model size not only for reducing the
computational and memory requirements~\cite{Cao14,Irschara,Li10,Park} but
also for minimizing communication overheads for IoT systems
as these models need to be transmitted from a cloud server to 
mobile devices.
In addition, the model is updated with 
newly arrived images. 
In this paper, a model pool
is constructed for model update (See Section~\ref{sec:model_update}). 

The ego-positioning phase is carried out on mobile devices. 
The 2D-to-3D image matching and localization component matches
images from a camera with the corresponding 3D point
cloud model of this area (roughly localized by GPS) 
to estimate 3D positions.
Notice that there are two communication links in the proposed system. 
One is to download 3D point cloud models to a mobile device, or preloaded
when a fixed walking/driving route is given.
However, the size of each uncompressed 3D point model (ranging from 105.1 MB
MB to 12.8 MB in our experiments) is still a crucial problem (as numerous
models are required for real-world long-range trips).
The other link is to upload the images to a 
cloud server for model update, but not required frequently. 

\section{Vision-Based Positioning System}
\label{sec:system}
As explained in Section~\ref{sec:overview}, there are four main
components in our system which are described in the following
sections. 

\subsection{Image-Based Modeling}
\label{sec:modeling}
Image-based modeling aims to construct a 3D model from a
number of input images \cite{Debevec}. 
One of the well-known image-based
modeling systems is the Photo Tourism method~\cite{Snavely}, which
uses structure from motion to estimate camera poses and 3D
scene geometry from images. 
In this section, we use their method to build the 3D point cloud model.
After images from mobile devices in a local area are acquired, the SIFT
\cite{Lowe} feature descriptors between each pair of images are then matched
using an approximate nearest neighbors kd-tree. 
A fundamental matrix is then estimated using RANSAC to remove outliers, which violate the geometric consistence. 
%

\begin{figure}[t]
\begin{center}
\includegraphics[width=0.8\linewidth]{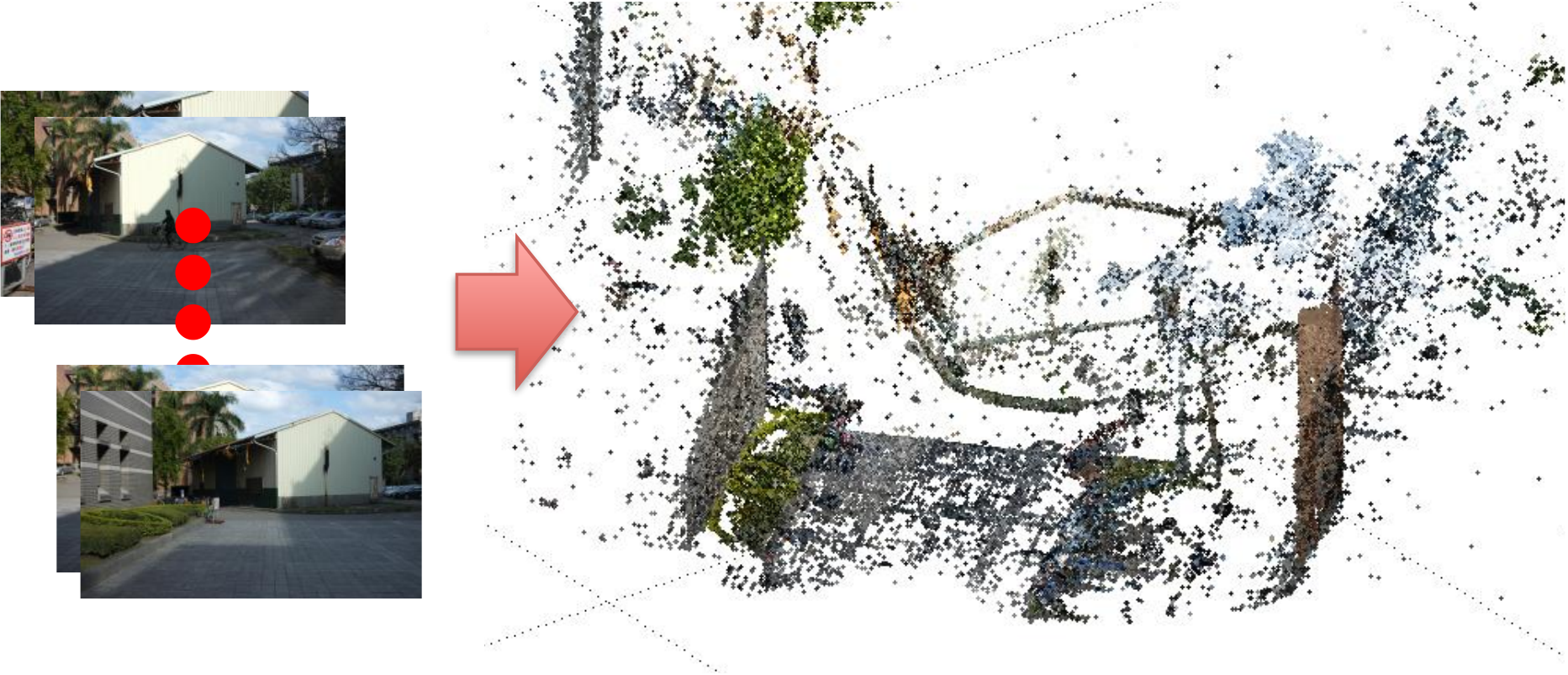}
\end{center}
   \caption{An example of image-based modeling from 400 images.}
\label{fig:ExampleModeling}
\end{figure}

Next, an incremental structure from motion method is
used to avoid bad local minimal solutions and to reduce the
computational load. 
It recovers camera parameters and 3D locations of feature points by
minimizing the sum of distances between the projections of 3D feature
points and their corresponding image features based on the following
objective function: 
\begin{equation}\label{equation1}
\min_{c_j,P_i} \sum_{i=1}^{n}\sum_{j=1}^{m}v_{ij}d\left ( Q\left (
    c_j,P_i \right),p_{ij} \right ), 
\end{equation}
where $c_j$ is the camera parameters of image $j$; 
$m$ is the number of images; 
$P_i$ is 3D coordinates of feature point $i$;
$n$ is the number of feature points;
$v_{ij}$ denotes the binary variables that equals 1 if point $i$ is
visible in image $j$ and 0 otherwise;
$Q(c_j, P_i)$ projects the 3D point $i$ onto the image $j$;
$p_{ij}$ is the corresponding image feature of $i$ on $j$;
and $d(.)$ is the distance function. 
This objective function can be solved by using bundle
adjustment,
and a 3D point cloud model $\textbf{P}$ is built. It includes the positions of 3D feature points and the corresponding SIFT feature descriptor list for each point.
An example of image-based modeling is 
shown in Figure~\ref{fig:ExampleModeling}. 


\subsection{Structure Preserving Model Compression}
\label{sec:model_compression}

How to condense the model so that it retains the ego-positioning capability of the full model is a central issue to reduce the computation, memory, and communication overheads.
A feasible way would be sorting the 3D points by their visibility and keep the most observed points.
However, it might cause non-uniform spatial distribution of the points because the points visible to many views could be distributed in a small area in an image.
Positioning with these points thus leads to an inaccurate estimation of the camera pose.
Following this idea, previous works \cite{Irschara,Li10,Park} formulated the point-selection problem as a \emph{set k-cover problem} to find a minimum set of 3D points that ensures at least $k$ points visible in each camera view.
However, they suffer easily from the problem of non-uniform spatial distribution with each view as discussed above.
The problem becomes particularly serious with the local area models constructed in our system.
To overcome this problem, we propose an approach that solves a \emph{weighted set k-cover problem}, where the weights are given to ensure the selected points to be fairly distributed on all planes and lines in the area.
The condensed model follows better the spatial structure of the scene.
It is crucial for our system to achieve sub-meter accuracy on session data.
Details of our algorithm are given in the following.

\begin{figure}[t]
\begin{center}
\subfigure[]{
\includegraphics[width=0.45\linewidth]{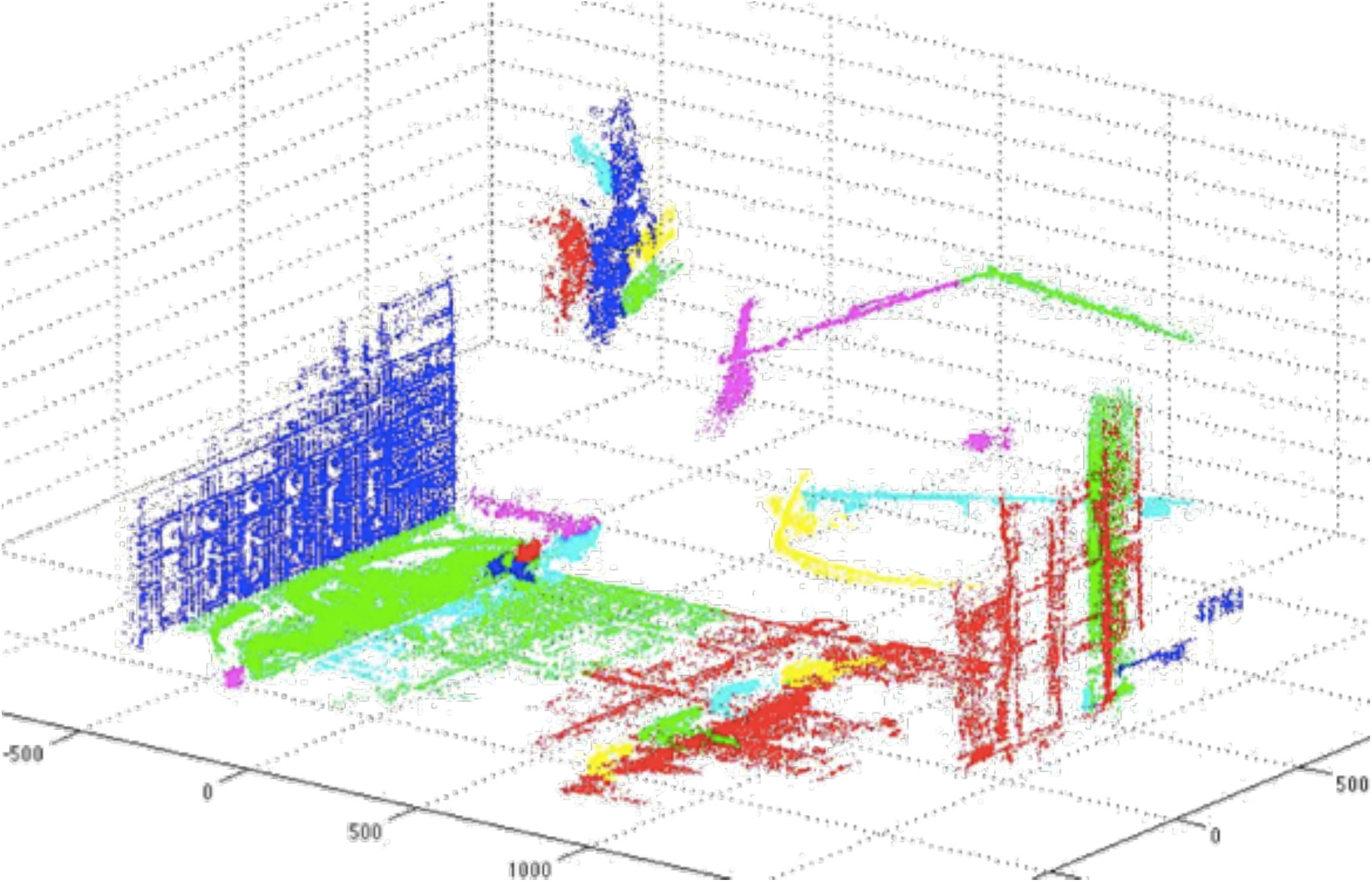}}
\subfigure[]{
\includegraphics[width=0.45\linewidth]{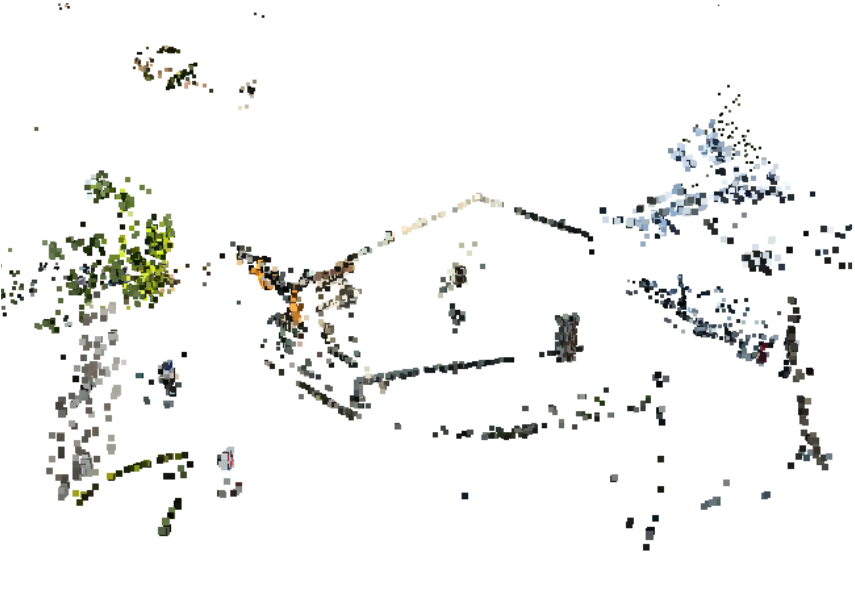}}
\end{center}
   \caption{(a) An example of plane and line detection. (b) An example of point cloud model after model compression.}
\label{fig:ExampleCompression}
\end{figure}

Let the 3D point cloud reconstructed be $\textbf{P}$ = \{$P_1$,...,$P_N$\}, $P_i \in \mathbb{R}^3$ and $N$ is the total number of points.
First, we detect planes and lines from the 3D point cloud by using RANSAC algorithm \cite{Yang}.
This procedure select three random points to generate a plane model and evaluate it by counting the number of points consistent to that model (\ie with the point-to-plane distance smaller than a threshold).
Repeat the random-selection procedures; then the plane having the most counts is selected as the detected plane.
We simply follow the one-at-of-time principle for detecting multiple planes.
After one plane is detected via RANSAC, the 3D points belonging to this plane are omitted, and then next plane will be detected from the remaining points.
This is repeated until that there are no planes with sufficient consistent counts.
Line detection is performed after plane detection in a similar process, unless two points are selected to generate a line-model hypothesis in each iteration.
An example of plane and line detections of the model in Figure~\ref{fig:ExampleModeling} is shown in Figure~\ref{fig:ExampleCompression}(a).

Note that, by using RANSAC and the one-at-a-time strategy, the planes and lines detected are usually distributed over the scene because the main structure (in terms of planes or lines) will be found at first.
This is helpful in selecting the non-local common-view points for condensing the 3D point model.
The planes and lines detected in the original point cloud model are denoted as $r_l$ $(l=1\cdots L)$, where $L$ is the total number of planes and lines.
The rest points not belonging to any plane or line is grouped as a single category $L+1$.
Given a 3D point $P_i$ in the 3D model reconstructed, we initially set its weight $w_i$ as
\begin{equation}
\label{equation_weight}
      w_i = \begin{cases}
		\sigma_l, & \text{ if } P_i\in r_l, \\
		\sigma_{L+1}, & \text{ otherwise },
		\end{cases}
\end{equation}
with
\begin{equation}
\label{equation_cal_ratio_l}
	\sigma _{l}=\left | \left \{ P_i|P_i\in r_l,\forall i \right \} \right |/N,
\end{equation}
\begin{equation}
\label{equation_cal_ratio_L+1}
	\sigma _{L+1}=\left | \left \{ P_i|P_i\notin r_l,\forall i,\forall l \right \} \right |/N.
\end{equation}
That is, the plane or line having a larger portion of points is given with a higher weight for point selection.

Based on the weights given in (\ref{equation_weight}), a weighted set k-cover problem is formulated.
However, the minimum set k-cover problem is a NP-hard.
Therefore, we use a greedy algorithm to solve it efficiently.
Let \{$c_1$,...,$c_m$\} be the $m$ camera views taken for the 3D model reconstruction in Section~\ref{sec:modeling}.
We hope to select a subset of points from $\textbf{P}$ such that there are at least $k$ points viewable for each camera, and the sum of weights of the selected points is maximized.

Following the greedy principle for solving a set covering problem, our approach iteratively selects the most visible point until at least $k$ points are selected for every camera.
In the beginning, we find one point from $\textbf{P}$ at first.
This point (denoted as $P_s$) satisfies the following criterion:
\begin{equation}
P_s = \mathop{\rm argmax}\limits_{P\in \textbf{P}} \sum_{j=1}^{m}w_i v(P,c_j),
\label{eq:onept}
\end{equation}
with
\begin{equation}
v(P,c_j) = \begin{cases}
		1, & \text{ if } P \in c_j, \\
		0, & \text{ otherwise }, 
		\end{cases}
\end{equation}
being the visibility of the point $P$ from the $j$-th camera.
Hence, $P_s$ found by (\ref{eq:onept}) is the most commonly visible point in terms of the respective weight.
We then remove $P_s$ from $\textbf{P}$ with $\textbf{P} \leftarrow \textbf{P}\backslash P_s$ in our greedy approach.
Then, our approach keeps finding the most visible point ($P_s$) based on the new point set $\textbf{P}$, and the procedure is iterated accordingly.

\noindent \textbf{Adaptive weight set k-cover}: In the above we assume that the weights $\{w_i\}$ are fixed.
However, although the planes and lines found by our approach are usually distributed over the scene, the points selected are still possibly centralized in a local region inside a single plane or line.
Therefore, we propose to adapt the weights in every iteration of our greedy algorithm.
Once the point $P_s$ is selected in an iteration, we reduce the selection possibility of the plane or line containing this point:
If $P_s \in r_l$, then $w_i$ will be divided by 2 for all points $P_i \in r_l$ in the next iteration.
In this way, the plane or line from which the points have already been selected will be weight-reduced, and the other planes or lines will have higher possibilities to be selected later.
A detailed algorithm can be found in Algorithm 1.
An example of model compression is shown in Figure~\ref{fig:ExampleCompression}(b). 
It shows that the main structure is kept and most of the noisy points are removed, and the model size is reduced from 105.1 MB to 14.4 MB.



\begin{algorithm}[t]
\label{algo:model_compression}
  \caption{Structure preserving model compression}
  \begin{algorithmic}[1]
    \Require 3D point cloud model $\textbf{P}$ = \{$P_1$,...,$P_N$\}, cameras \{$c_1$,...,$c_m$\}, integer $k$
    \State Initialize the compressed model $\textbf{M} \leftarrow \emptyset$, number of covered points $C[j] = 0$, for all camera $c_j$
    \State Detect planes and lines {$r_1$,...,$r_L$} in \textbf{P} by RANSAC algorithm
    \State Assign weight $w_i$ to each point $P_i$ by (\ref{equation_weight})
    \While{$C[j]$ $<$ $k$, for all camera $c_j$}
      \State Select $P_s$ by (\ref{eq:onept})
      \State $\textbf{M} \leftarrow P_s$
      \State $\textbf{P} \leftarrow \textbf{P}\backslash P_s$
      \ForAll {cameras $c_j$}
      	\If{$P_s$ $\in$ $c_j$}
		\State $C[j]=C[j]+1$
	\EndIf;
      \EndFor;
      \ForAll {3D points $P_i$}
      	\If{$P_s$ $\in$ $r_l$ and $P_i$ $\in$ $r_l$}
		\State $w_i = w_i/2$
	\EndIf;
	\If{$P_s$ $\notin$ $c_j$ for all $j$ s.t. $C[j] < k$}
		\State $w_i = 0$
	\EndIf;
      \EndFor;
      \State Normalize $w_i$
    \EndWhile;
  \State \Return compressed model $\textbf{M}$
  \end{algorithmic}
\end{algorithm}

Because the 3D point cloud model is constructed in advance, a visual word tree can be built for speedup before the image matching procedure.
A visual word is a cluster of similar feature descriptors and the visual words are obtained by k-means clustering.
Afterwards, a kd-tree \cite{Lowe} is built based on FLANN library \cite{Muja} for fast indexing \cite{Sattler11}.

\subsection{2D-to-3D Image Matching and Localization}

\begin{figure}[ht]
\begin{center}
\includegraphics[width=0.9\linewidth]{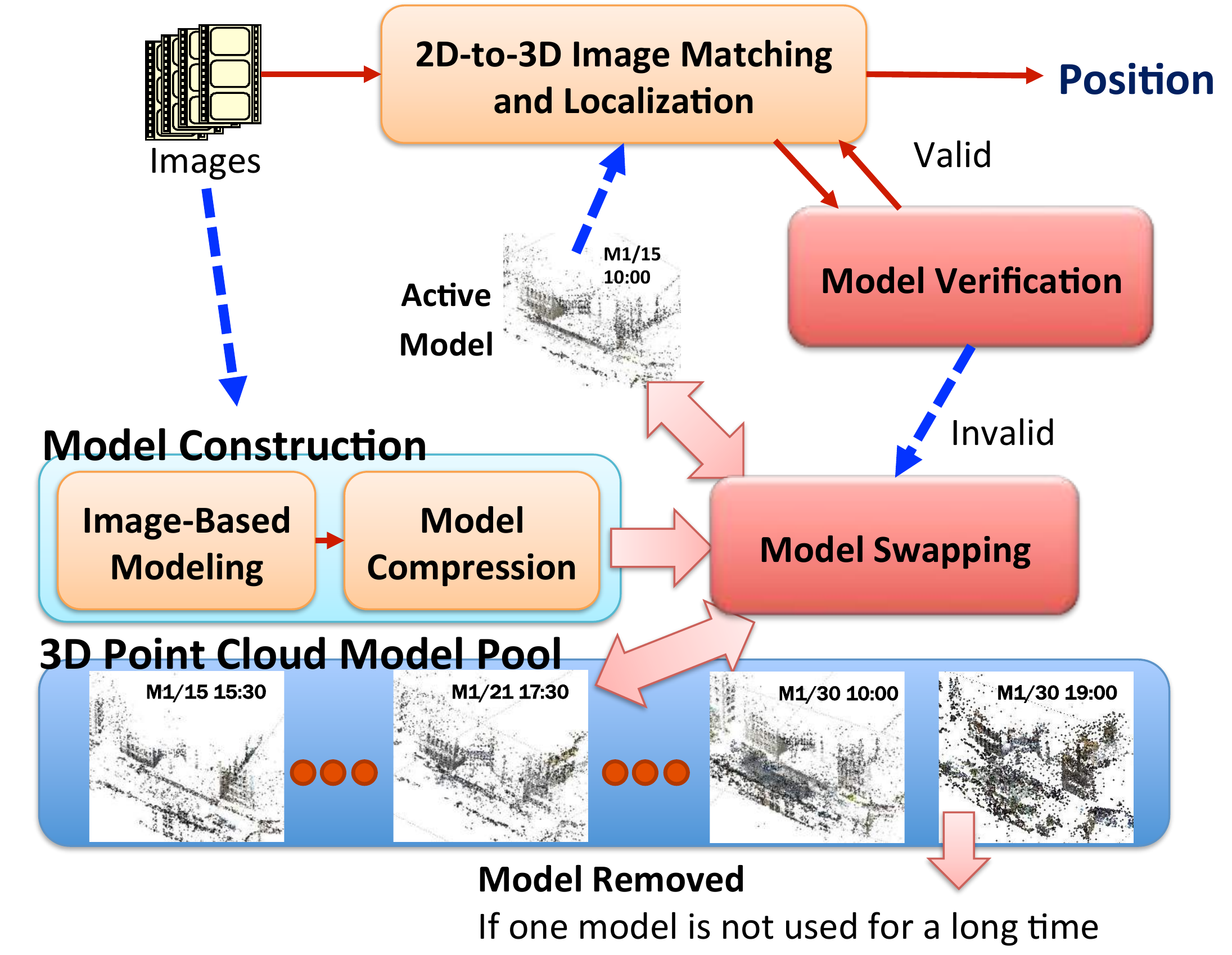}
\end{center}
   \caption{An overview of our model update algorithm.}
\label{fig:OverviewUpdate}
\end{figure}

Given a test image, the interesting points (2D) are detected and their SIFT descriptors are computed.
2D-to-3D matching is referred to as finding the correspondence of the 2D points in the test image and the 3D points in the compressed model.
Then, the camera position can be estimated based on the correspondence.
In our approach, a 2D-to-3D correspondence will be accepted if the first and second nearest neighbors in the descriptor space pass the ratio test with a threshold (that is set as 0.7 in our system).
To speed it up, a prioritized correspondence search \cite{Sattler11} is applied. 
After finding correspondences $\{\{p_1,P_1\},...,\{p_{N_c}, P_{N_c}\}\}$, where $p$ is the corresponding 2D feature point of $P$ and $N_c$ is number of correspondences, we run the 6-point DLT algorithm \cite{Hartley} with RANSAC for ego-positioning of the camera.

\begin{figure*}[ht]
\begin{center}
\includegraphics[width=0.8\linewidth]{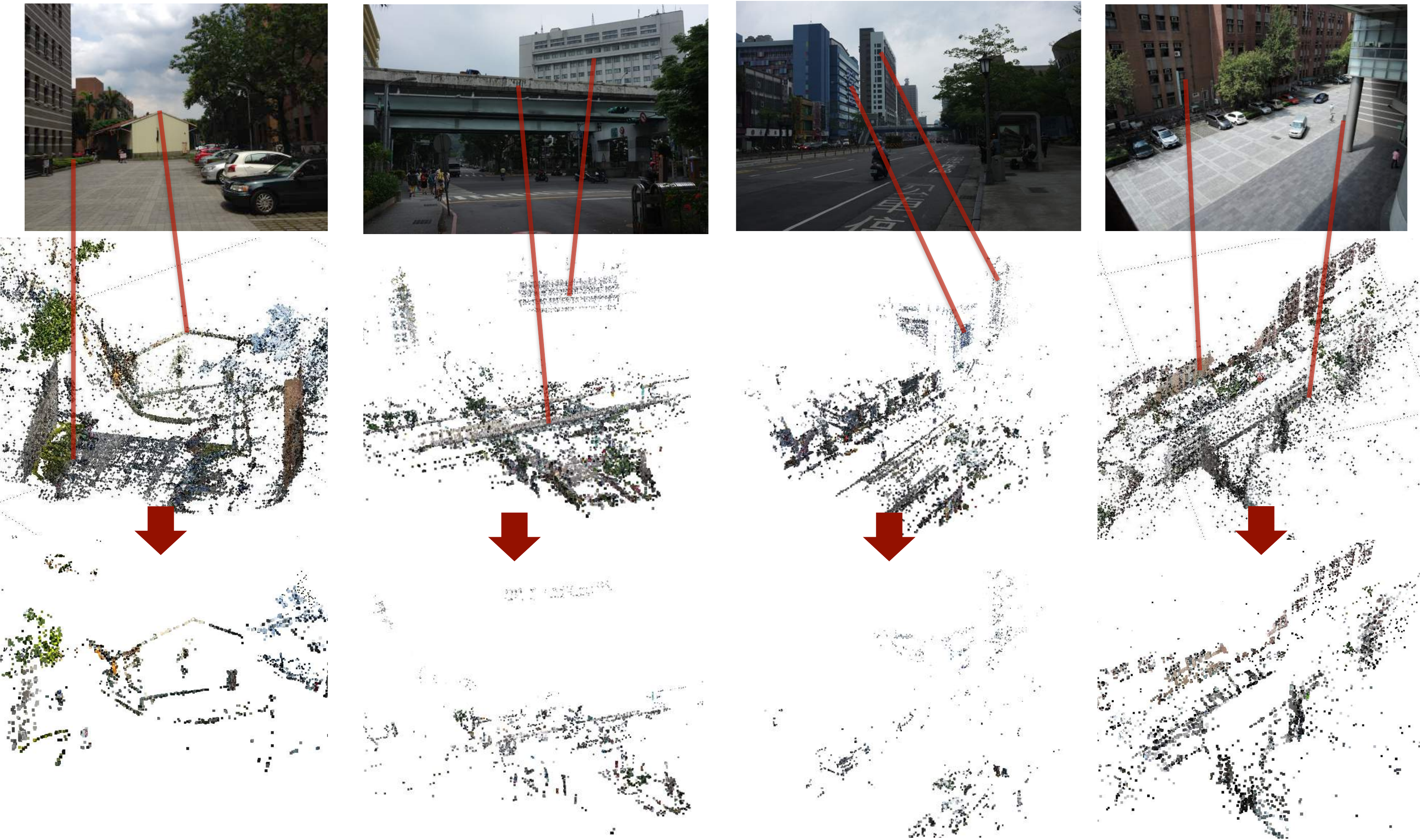}
\end{center}
   \caption{Four scenes: Scene \#1, Scene \#2, Scene \#3, and Scene \#4 from left to right respectively. The first row is the images of the scenes. The second row shows the constructed models, and the compressed models by our method are shown in the third row.}
\label{fig:FourScenes}
\end{figure*}

\subsection{Model Update}
\label{sec:model_update}

Because the scenes are not always unchanged in the outdoor
environments and all vision-based approaches are sensitive to
scene changes, how to update the model remains an issue. 
In practice, the lighting condition, weather, and even the structure of the scene would be
changed.
Previous works assume an up-to-date model having been constructed and do not handle the updating
problem.
We tackle this problem and an algorithm is proposed for solving it.

An overview is shown in Figure~\ref{fig:OverviewUpdate}.
Instead of using only one 3D point cloud model, a model pool is kept for every
local area in our system.
It contains multiple 3D point cloud models
of the same area constructed in different sessions of time.
Notice that, we do not merge more observations into a single model here, because it will lead to more ambiguous when matching features and further enlarge the model size. That is why we use multiple small models. Furthermore, only one active model is selected and will be transmitted in our model update method.
There are two main components included: model verification and model swapping.
In the ego-positioning phase, the model verification component verifies
whether the input image can be registered with the model well by the following
criterion,
\begin{equation}\label{equation_criterion}
         (N_c > T_1) \mbox{\ and \ } (N_I/N_c > T_2),
\end{equation}
where $N_c$ is number of correspondences and $N_I$ is the number of inlier correspondences found by the pose-estimation algorithm \cite{Hartley}.
$T_1$ and $T_2$ are leant from the training data and set as 50 and 50\%, respectively, in our experiments.

If multiple observations claim that the currently active model is invalid, the model swapping component will be activated. 
It first evaluates all of the models in the model pool by (\ref{equation_criterion}) with the images collected in this session. 
Second, the model with the highest score (obtained based on (\ref{equation_criterion})) will then be selected as the new active model if its score is higher than a threshold.
Third, if all of the model scores are below the threshold, a new model will be constructed by the images collected in this session to replace the active model.
Fourth, if a model in the model pool is no longer used for a long time, it will be removed to avoid the unlimited growth of the model pool size.

We have shown a few models (8 models for daytime in our experiments) are sufficient to represent one scene for positioning even when scene changes over a long period of time. It shows the case of constructing new model rarely happens and the proposed model swapping strategy works well for real applications.


\section{Experimental Results}
\label{sec:results}

In this section, we first evaluate the positioning of a single still image with both local and up-to-date models in four scenes.
Second, video sequences are tested.
It shows that larger deviation will be obtained than testing of the still images because of occlusion and/or motion blurs, but the results can be compensated by applying temporal smoothing.
Third, a dataset with more than 14,000 images of session data with 106 sessions is constructed and released.
Finally, the proposed model update algorithm is evaluated.

The ground truth positions in our experiments are all measured in real scenes manually.
Unlike most recent works \cite{Lim,Middelberg,Ventura} using the results generated by offline structure-from-motion algorithms as the ground truth, all test images have ground-truths in our setting.
The unregistered images will be skipped in the setup of the above works.
Furthermore, we have found a number of images fail of being registered in our experiments, particularly for those taken at night.

\begin{table}[t]
\caption{Positioning error (cm) of single still image and number of points and model size after model compression. Red font and blue font show the minimun and second minimum mean, stdev., or model sizes of four compression method, respectively.}
\label{table:ErrorImage}
\small
\begin{tabular}{|l|c|c|c|c|c|}
\hlinewd{1.5pt}
\multicolumn{2}{|c|}{Scene}                                                                             & \#1                                  & \#2                                  & \#3                                  & \#4                                  \\ \hline
                                                                                            & Mean      & 22.9                                 & 21.6                                 & 35.9                                 & 33.7                                 \\ \cline{2-6} 
                                                                                            & Stdev.    & 8.1                                  & 10.8                                 & 15.6                                 & 9.8                                  \\ \cline{2-6} 
                                                                                            & \# points & 187,572                              & 53,568                               & 33,190                               & 85,447                               \\ \cline{2-6} 
\multirow{-4}{*}{\begin{tabular}[c]{@{}l@{}}Original\\ Model\end{tabular}}                  & Size(MB) & 105.1                                & 21.5                                 & 12.8                                 & 26.4                                 \\ \hlinewd{1.5pt}
                                                                                            & Mean      & 68.7                                 & 38.2                                 & {\color[HTML]{FE0000} \textbf{30.9}} & {\color[HTML]{00009B} 40.0}          \\ \cline{2-6} 
                                                                                            & Stdev.    & 143.5                                & 49.2                                 & 40.5                                 & {\color[HTML]{00009B} 38.3}          \\ \cline{2-6} 
                                                                                            & \# points & 8,397                                & 4,329                                & 3,268                                & 8,263                                \\ \cline{2-6} 
\multirow{-4}{*}{\begin{tabular}[c]{@{}l@{}}Compressed\\ by baseline\\ method\end{tabular}} & Size(MB) & {\color[HTML]{FE0000} \textbf{10.2}}                                 & {\color[HTML]{00009B} 2.5}                                  & {\color[HTML]{00009B} 1.5}                                  & 5.4                                  \\ \hlinewd{1.5pt}
                                                                                            & Mean      & {\color[HTML]{FE0000} \textbf{21.3}} & 38.2                                 & 46.24                                & 42.9                                 \\ \cline{2-6} 
                                                                                            & Stdev.    & {\color[HTML]{FE0000} \textbf{7.9}}  & 20.1                                 & 27.1                                 & 72.8                                 \\ \cline{2-6} 
                                                                                            & \# points & 8,580                                & 5,101                                & 3,345                                & 8,272                                \\ \cline{2-6} 
\multirow{-4}{*}{\begin{tabular}[c]{@{}l@{}}Compressed\\ by \cite{Li10}\end{tabular}}     & Size(MB) & 17.8                                 & 3.0                                  & 1.6                                  & {\color[HTML]{00009B} 4.8}                                  \\ \hlinewd{1.5pt}
                                                                                            & Mean      & {\color[HTML]{00009B} 24.7}          & {\color[HTML]{00009B} 27.7}          & 41.1                                 & 40.1                                 \\ \cline{2-6} 
                                                                                            & Stdev.    & 19.4                                 & {\color[HTML]{00009B} 17.5}          & {\color[HTML]{00009B} 23.5}          & 48.5                                 \\ \cline{2-6} 
                                                                                            & \# points & 8,519                                & 4,557                                & 3,149                                & 8,166                                \\ \cline{2-6} 
\multirow{-4}{*}{\begin{tabular}[c]{@{}l@{}}Compressed\\ by \cite{Cao14}\end{tabular}}    & Size(MB) & 19.9                                 & 2.7                                  & 1.7                                  & 5.0                                  \\ \hlinewd{1.5pt}
                                                                                            & Mean      & 25.9                                 & {\color[HTML]{FE0000} \textbf{24.9}} & {\color[HTML]{00009B} 39.0}          & {\color[HTML]{FE0000} \textbf{33.2}} \\ \cline{2-6} 
                                                                                            & Stdev.    & {\color[HTML]{00009B} 11.7}          & {\color[HTML]{FE0000} \textbf{12.9}} & {\color[HTML]{FE0000} \textbf{21.8}} & {\color[HTML]{FE0000} \textbf{14.0}} \\ \cline{2-6} 
                                                                                            & \# points & 7,781                                & 4,351                                & 3,228                                & 8,196                                \\ \cline{2-6} 
\multirow{-4}{*}{\begin{tabular}[c]{@{}l@{}}Compressed\\ by\\ our method\end{tabular}}        & Size(MB) & {\color[HTML]{00009B} 14.4}                                 & {\color[HTML]{FE0000} \textbf{2.2}}                                  & {\color[HTML]{FE0000} \textbf{1.5}}                                 & {\color[HTML]{FE0000} \textbf{4.3}}                                  \\ \hlinewd{1.5pt}
\end{tabular}
\end{table}

\subsection{Positioning Evaluation of Single Still Image}

\begin{figure}[t]
\begin{center}
\includegraphics[width=0.8\linewidth]{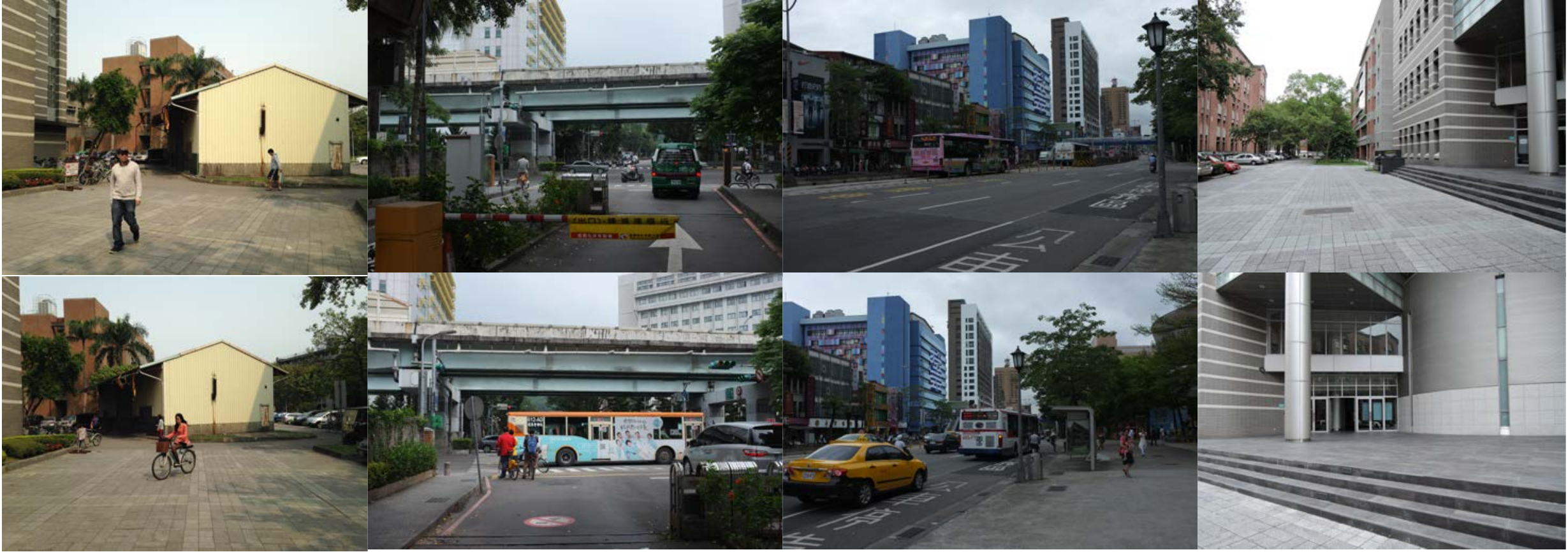}
\end{center}
   \caption{Some examples of testing images in four scenes.}
\label{fig:ExampleTest}
\end{figure}

To show that a local and up-to-date model can lead to sub-meter positioning accuracy, we test the positioning algorithm in four scenes. Figure~\ref{fig:FourScenes} shows the scenes, constructed models, and the compressed models generated by our algorithm.
The models are reconstructed with 400, 179, 146, and 341 training images, respectively.
The test images are taken in the same session of the training images. 
There are 50, 24, 28, and 88 test images, respectively, and Figure~\ref{fig:ExampleTest} shows some of them. 
Table~\ref{table:ErrorImage} demonstrates the positioning results and the comparison of model compressions with different methods, where the baseline method is to keep 5\% (for scene \#1) or 10\% (for other scenes) mostly seen points on each plane or line, Li \etal method \cite{Li10} is the approach to solve set k-cover problem ($k$ = 720, 500, 250, and 200 for each scene, respectively), Cao and Snavely method \cite{Cao14} solves a probabilistic k-cover problem ($k$ = 430, 300, 250, and 200 for each scene, respectively), and our method is to formulate a weighted set k-cover problem ($k$ = 500, 300, 370, and 185 for each scene, respectively) as depicted in Section~\ref{sec:model_compression}, where we select a proper $k$ for each method to ensure the number of points after model compression being similar to each other for fair comparison.

\begin{figure}[t]
\begin{center}
\subfigure[]{
\includegraphics[width=0.4\linewidth]{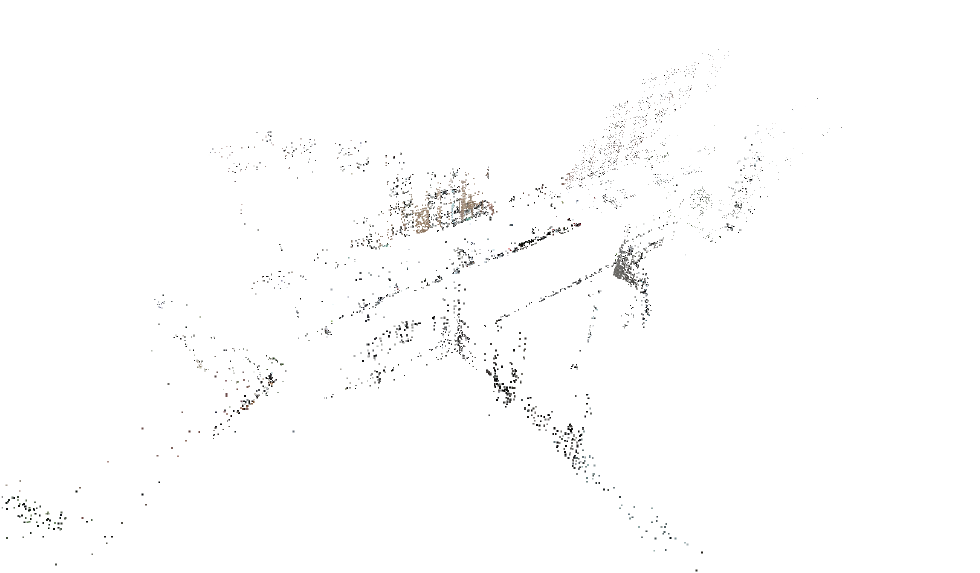}}
\subfigure[]{
\includegraphics[width=0.4\linewidth]{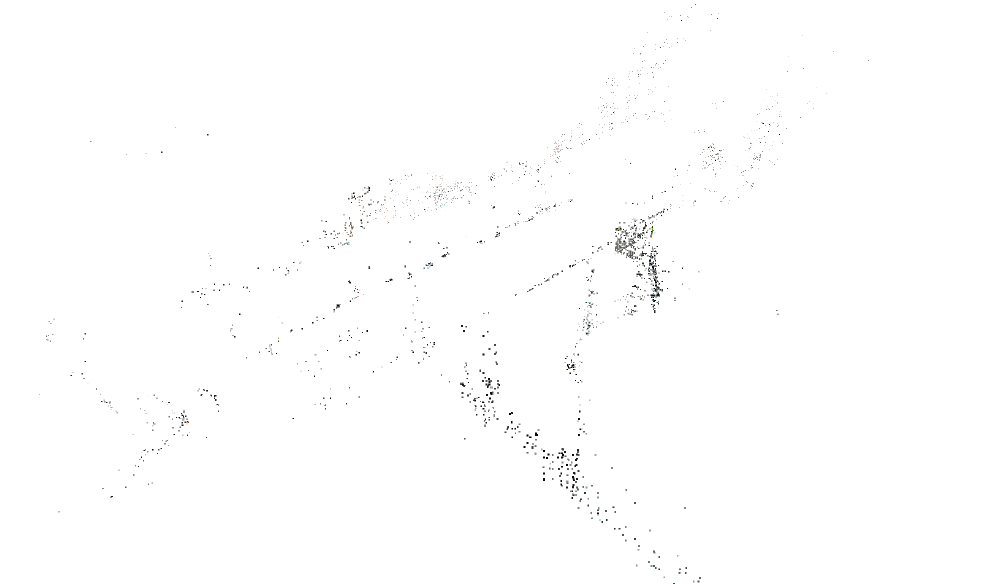}}
\subfigure[]{
\includegraphics[width=0.4\linewidth]{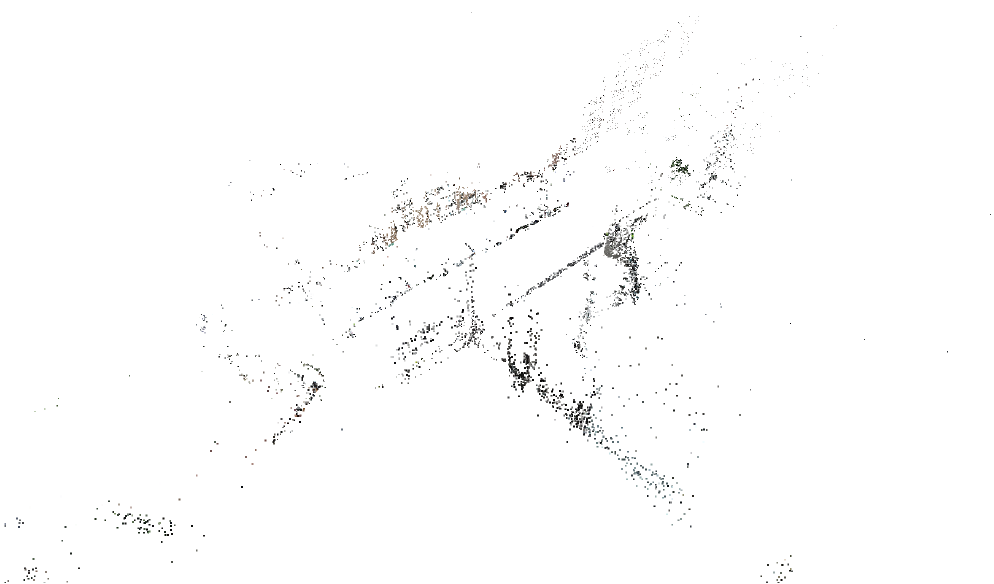}}
\subfigure[]{
\includegraphics[width=0.4\linewidth]{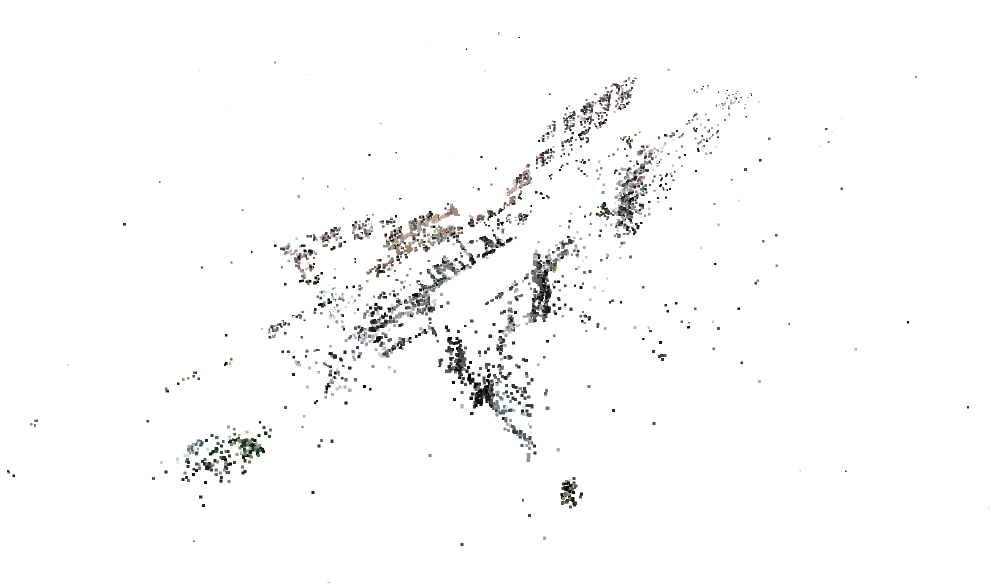}}
\end{center}
   \caption{The model compression results of (a) baseline method, (b) Li's method \cite{Li10}, (c) Cao's method \cite{Cao14} and (d) our method, in scene \#4.}
\label{fig:CompressionResults}
\end{figure}

It shows that sub-meter accuracy can be achieved even when model compression is performed for a local and up-to-date model. 
Furthermore, our compression method performs favorably against other methods in both positioning accuracy and model size reduction because of its capability of preserving the structure information better, which would lead to more stable positioning results. Table~\ref{table:ErrorImage} shows our method performs slightly worse than \cite{Cao14,Li10} in Scene \#1, but it reduces model size more and further leads to the best performances in other scenes. Notice that both methods \cite{Cao14,Li10} perform even worse than baseline method in Scene \#4, and all of them are with large standard deviation (about three times larger than ours). Moreover, our method outperforms than other methods in model size reduction, too. Figure~\ref{fig:CompressionResults} shows the compressed models of scene \#4, and our method keeps more structural information compared with other methods.  

We do not focus on real-time processing in this paper. The average execution time of one image with 900x600 resolutions running on a PC with Intel Core i7-4770K processor and 16G ram is 1.4089 seconds (including SIFT feature extraction 1.274 seconds, feature matching 0.1319 seconds, and RANSAC pose estimation 0.0003 seconds).
Most of the time are spent on SIFT feature extraction and can be improved by GPU or multi-thread implementations in the future.

\subsection{Positioning Evaluation of Video Sequence}

\begin{figure}[t]
\begin{center}
\subfigure[]{
\includegraphics[width=0.2\linewidth]{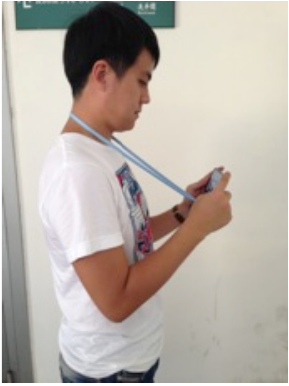}}
\subfigure[]{
\includegraphics[width=0.7\linewidth]{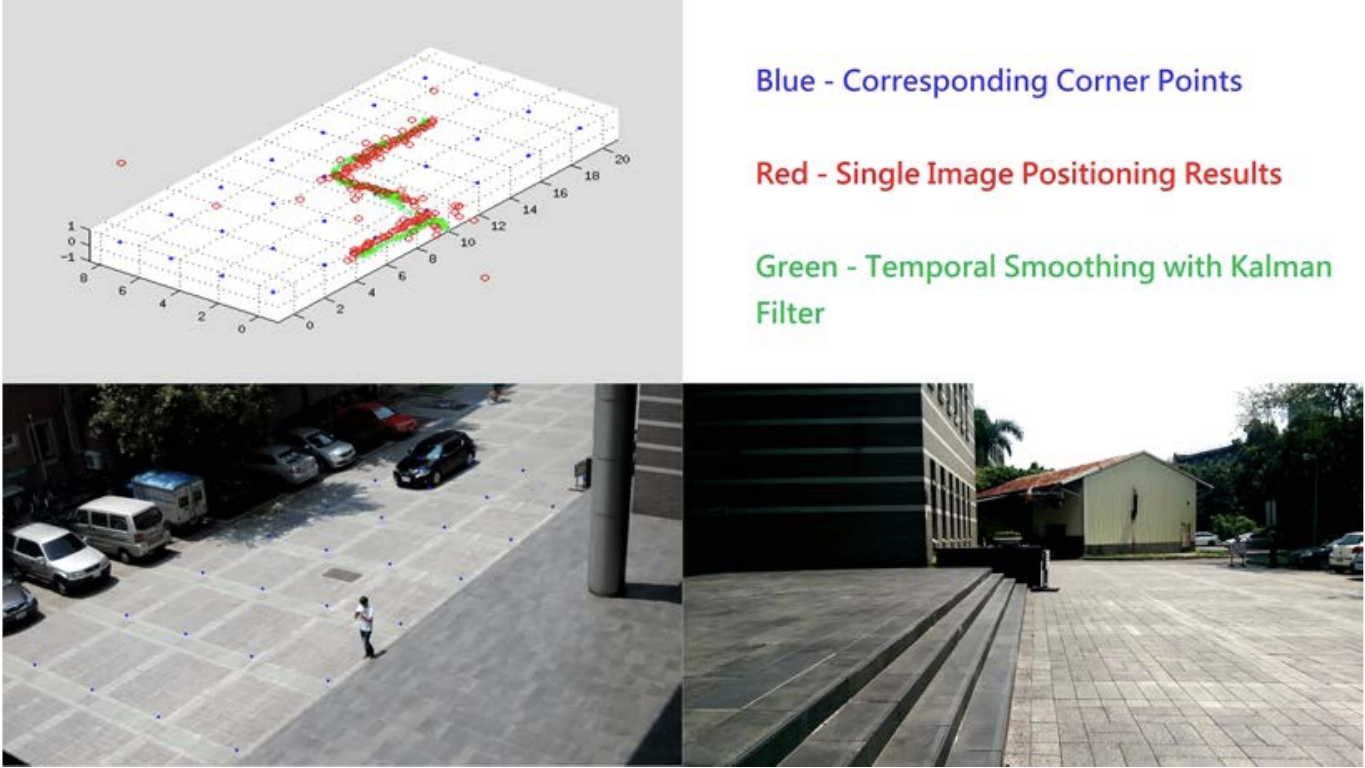}}
\end{center}
   \caption{(a) The setup of taking video sequences on human. (b) An example of results of video evaluation. Bottom-right: image from smartphone; bottom-left: image from camera on third floor; upper-left: positioning results.}
\label{fig:ExampleVideo}
\end{figure}

We then test two video sequences in Scene \#4. For each test video, we recored the videos from two cameras with 10 frames per second (fps). One is taken by a camera on the third floor of a nearby building, as shown in the bottom-left image of Figure~\ref{fig:ExampleVideo}(b). This is used for display and providing visual verification. 
We click the coordinate of the man's foot on the ground plane in each image and then transform the coordinates to the ground plane as the ground truth position. 
The other is taken by a smart phone as shown in \ref{fig:ExampleVideo}(a) and the bottom-right image of Figure~\ref{fig:ExampleVideo}(b), which is used for vision-based positioning. The upper-left image in Figure~\ref{fig:ExampleVideo}(b) demonstrates the estimation results, where red and green circles are the positioning results with still images and the results after temporal smoothing by Kalman filter \cite{Kalman}, respectively. 
More details can be seen in the demo video: PositioningDemo.mp4. 
Table~\ref{table:ErrorVideo} shows the quantitative results. Due to occlusions or motion blurs in the video, there are sometimes outlier or noisy estimations by using still-image positioning, but they can be compensated well after temporal smoothing.

\begin{figure}[t]
\begin{center}
\includegraphics[width=0.8\linewidth]{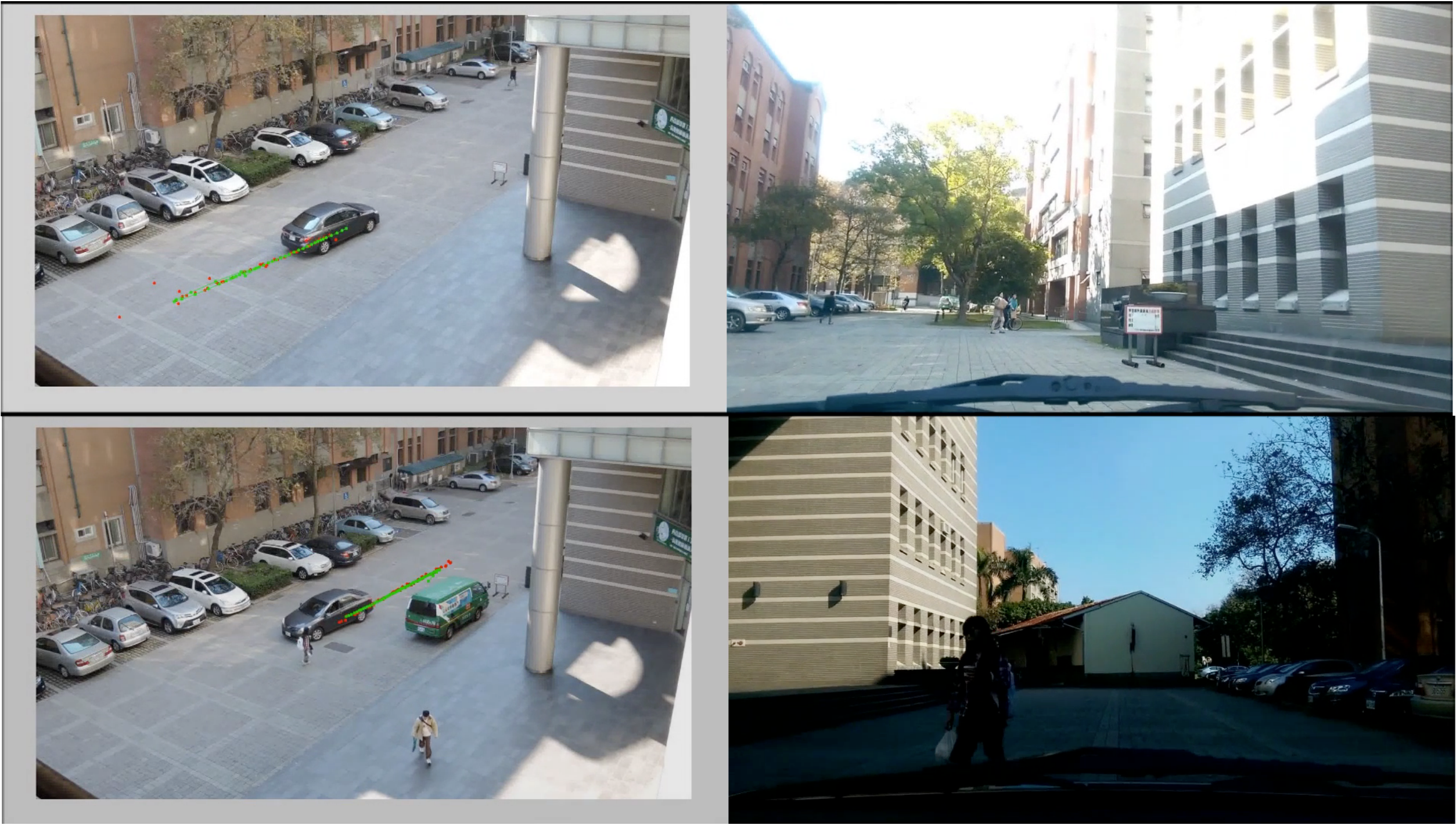}
\end{center}
   \caption{Examples of results of video evaluation. Right column: image from dash camera; left column: image from camera on third floor,  where red and green circles are the positioning results with still images and the results after temporal smoothing by Kalman filter.}
\label{fig:CarVideo}
\end{figure}

We also have another evaluation of video sequences taken by a dash camera on a car. Figure~\ref{fig:CarVideo} shows the results and more results can be seen in the demo video: PositioningDemo.mp4. Although we do not have quantitative results for these video sequences, because it is hard to determine the ground truth positions of the dash camera, the demo video shows high accuracies are achieved while the estimated positions lying on the car well. Notice that the test video sequences were taken on different days and model update algorithm have been applied to these video evaluation. More results of model update will be shown in Section~\ref{sec:update_results}.

\subsection{Long-Term Positioning Dataset}

\begin{figure*}[t]
\begin{center}
\subfigure[]{
\includegraphics[width=0.8\linewidth]{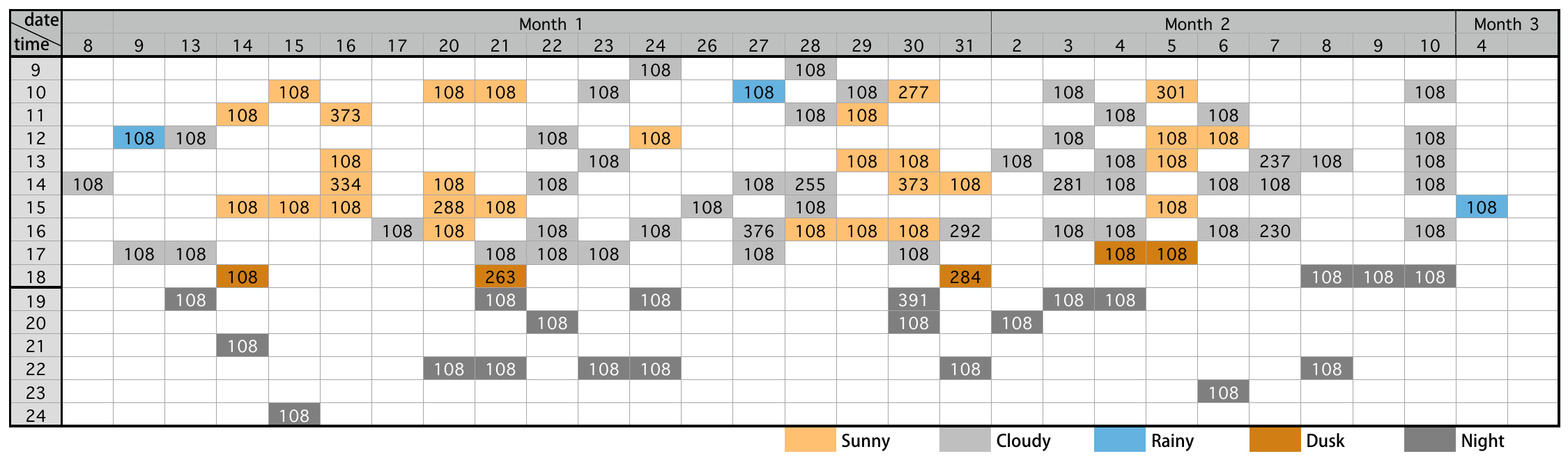}}
\subfigure[]{
\includegraphics[width=0.8\linewidth]{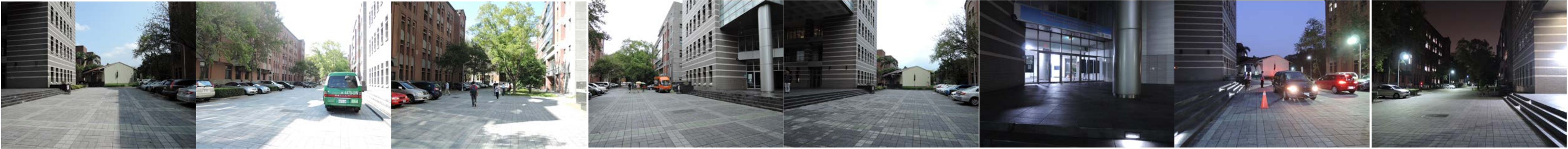}}
\end{center}
   \caption{The long-term positioning dataset: (a) Session distribution with the number of images, and (b) Some examples of images.}
\label{fig:Dataset}
\end{figure*}

To evaluate the model update, a dataset with long-term session data is built.
It contains more than 1 month, 106 sessions, 14,275 images, and 9,720 of them are with manually measured ground truth positions. There is no positioning dataset with such session data for evaluation, to our best knowledge.
It includes the situations of sunny, cloudy, rainy, night, and so on. 
The session distribution and some sample images are shown in Figure~\ref{fig:Dataset}. 

\begin{table}[t]
\caption{Positioning error (cm) of video sequences.}
\label{table:ErrorVideo}
\small
\centering
\begin{tabular}{|c|c|c|c|}
\hlinewd{1.5pt}
\multicolumn{2}{|c|}{Video sequence}         & \#1  & \#2   \\ \hline
\multicolumn{2}{|c|}{Frame number}           & 200  & 281   \\ \hlinewd{1.5pt}
\multirow{2}{*}{Single still image} & Mean   & 60.1 & 88.4  \\ \cline{2-4} 
                                    & Stdev. & 53.8 & 116.7 \\ \hlinewd{1.5pt}
\multirow{2}{*}{Temporal smoothing} & Mean   & 37.2 & 41.8  \\ \cline{2-4} 
                                    & Stdev. & 18.3 &  26.3     \\ \hlinewd{1.5pt}
\end{tabular}
\end{table}

\begin{figure*}[ht]
\begin{center}
\includegraphics[width=1\linewidth]{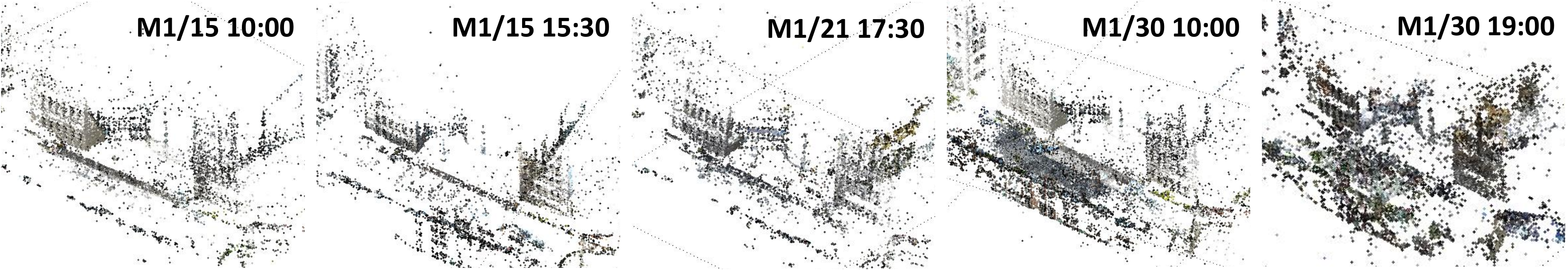}
\end{center}
   \caption{Some models in the model pool after learning.}
\label{fig:ModelPool}
\end{figure*}

\begin{table*}[t]
\caption{Positioning error (cm) of the session data in other months.}
\label{table:ErrorUpdate}
\small
\centering
\begin{tabular}{|c|c|c|c|c|c|c|c|c|c|c|}
\hlinewd{1.5pt}
\multicolumn{2}{|c|}{Test date \& time}                                                                                                                       & \begin{tabular}[c]{@{}c@{}}M2/3\\ 10:30\end{tabular}           & \begin{tabular}[c]{@{}c@{}}M2/4\\ 17:30\end{tabular}          & \begin{tabular}[c]{@{}c@{}}M2/5\\ 15:00\end{tabular}          & \begin{tabular}[c]{@{}c@{}}M2/6\\ 12:00\end{tabular}          & \begin{tabular}[c]{@{}c@{}}M2/7\\ 14:00\end{tabular}           & {\color[HTML]{00009B} \textbf{\begin{tabular}[c]{@{}c@{}}M2/8\\ 22:30\end{tabular}}}          & {\color[HTML]{00009B} \textbf{\begin{tabular}[c]{@{}c@{}}M2/9\\ 18:30\end{tabular}}}          & \begin{tabular}[c]{@{}c@{}}M2/10\\ 10:00\end{tabular}          & {\color[HTML]{FE0000} \textbf{\begin{tabular}[c]{@{}c@{}}M3/4\\ 14:30\end{tabular}}}           \\ \hlinewd{1.5pt}
\multicolumn{2}{|c|}{Weather}                                                                                                                                 & Cloudy                                                         & Dusk                                                          & Sunny                                                         & Sunny                                                         & Cloudy                                                         & {\color[HTML]{00009B} \textbf{Night}}                                                         & {\color[HTML]{00009B} \textbf{Night}}                                                         & Cloudy                                                         & {\color[HTML]{FE0000} \textbf{Rainy}}                                                          \\ \hlinewd{1.5pt}
                                                                                      & Mean                                                                  & 72.4                                                           & NaN                                                           & 74.2                                                          & 171.4                                                         & 91.6                                                           & {\color[HTML]{00009B} \textbf{1136}}                                                          & {\color[HTML]{00009B} \textbf{NaN}}                                                           & 135.9                                                          & {\color[HTML]{FE0000} \textbf{4107.8}}                                                         \\ \cline{2-11} 
\multirow{-2}{*}{\begin{tabular}[c]{@{}c@{}}Fixed model\\ (M1/20 15:30)\end{tabular}} & Stdev.                                                                & 40.4                                                           & NaN                                                           & 42.5                                                          & 318.6                                                         & 174.7                                                          & {\color[HTML]{00009B} \textbf{850}}                                                           & {\color[HTML]{00009B} \textbf{NaN}}                                                           & 180.2                                                          & {\color[HTML]{FE0000} \textbf{1264.7}}                                                         \\ \hlinewd{1.5pt}
                                                                                      & Mean                                                                  & 30.6                                                           & 39.3                                                          & 33.7                                                          & 28.2                                                          & 30.8                                                           & {\color[HTML]{00009B} \textbf{441.5}}                                                         & {\color[HTML]{00009B} \textbf{34.2}}                                                          & 41.5                                                           & {\color[HTML]{FE0000} \textbf{36.2}}                                                           \\ \cline{2-11} 
                                                                                      & Stdev.                                                                & 12.5                                                           & 16.8                                                          & 12.8                                                          & 11.7                                                          & 14.3                                                           & {\color[HTML]{00009B} \textbf{352.1}}                                                         & {\color[HTML]{00009B} \textbf{28}}                                                            & 13.3                                                           & {\color[HTML]{FE0000} \textbf{15.8}}                                                           \\ \cline{2-11} 
\multirow{-3}{*}{\begin{tabular}[c]{@{}c@{}}Model\\ update\\ applied\end{tabular}}    & \begin{tabular}[c]{@{}c@{}}Selected\\ model\\ \& weather\end{tabular} & \begin{tabular}[c]{@{}c@{}}M1/27\\ 16:00\\ Cloudy\end{tabular} & \begin{tabular}[c]{@{}c@{}}M1/30\\ 19:00\\ Night\end{tabular} & \begin{tabular}[c]{@{}c@{}}M1/30\\ 14:30\\ Sunny\end{tabular} & \begin{tabular}[c]{@{}c@{}}M1/16\\ 11:00\\ Sunny\end{tabular} & \begin{tabular}[c]{@{}c@{}}M1/27\\ 16:00\\ Cloudy\end{tabular} & {\color[HTML]{00009B} \textbf{\begin{tabular}[c]{@{}c@{}}M1/30\\ 19:00\\ Night\end{tabular}}} & {\color[HTML]{00009B} \textbf{\begin{tabular}[c]{@{}c@{}}M1/30\\ 19:00\\ Night\end{tabular}}} & \begin{tabular}[c]{@{}c@{}}M1/27\\ 16:00\\ Cloudy\end{tabular} & {\color[HTML]{FE0000} \textbf{\begin{tabular}[c]{@{}c@{}}M1/27\\ 16:00\\ Cloudy\end{tabular}}} \\ \hlinewd{1.5pt}
\end{tabular}
\end{table*}

\subsection{Positioning Evaluation with Model Update}
\label{sec:update_results}

We then apply our model update algorithm to the long-term dataset. In the end of the first month, there are totally 21 distinct models built in the model pool, with 8 for daytime and 13 for night. 
Figure~\ref{fig:ModelPool} shows some of them. 
It shows that a few of models are sufficient for positioning in the daytime, but it is hard to register images at night. 
Table~\ref{table:ErrorUpdate} presents some of the evaluation results in other months. 
Note that all models are selected automatically with the first ten images in that session by using our method in Section~\ref{sec:model_update}. 
It shows that our method can select a suitable model, and the results will be much improved when model update is applied in comparison to the use of a fixed model, for example, there are 100 times improvement at M3/4 14:30 (red font in the table). The fixed model is even worse and can not be registered in some sessions (e.g. M2/4 17:30). Furthermore, we find positioning in the night scene is still challenging, which is not perfectly solved in this paper. The night scene (blue font in the table) with sufficient lighting, e.g. M2/9 18:30, is tackled well, but the performance degrades under poor lighting conditions (e.g. M2/8 22:30). More efforts will be paid for night scene positioning in the future.

\subsection{Positioning for Augmented Reality Display}
\label{sec:app_results}

\begin{figure}[t]
\begin{center}
\subfigure[]{
\includegraphics[width=0.48\linewidth]{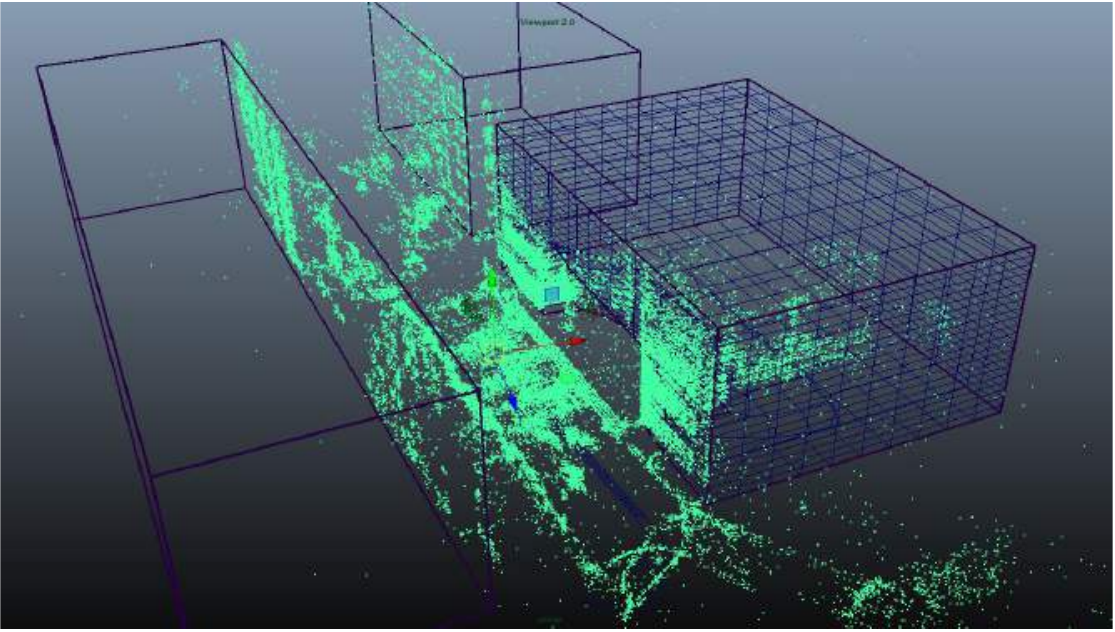}}
\subfigure[]{
\includegraphics[width=0.48\linewidth]{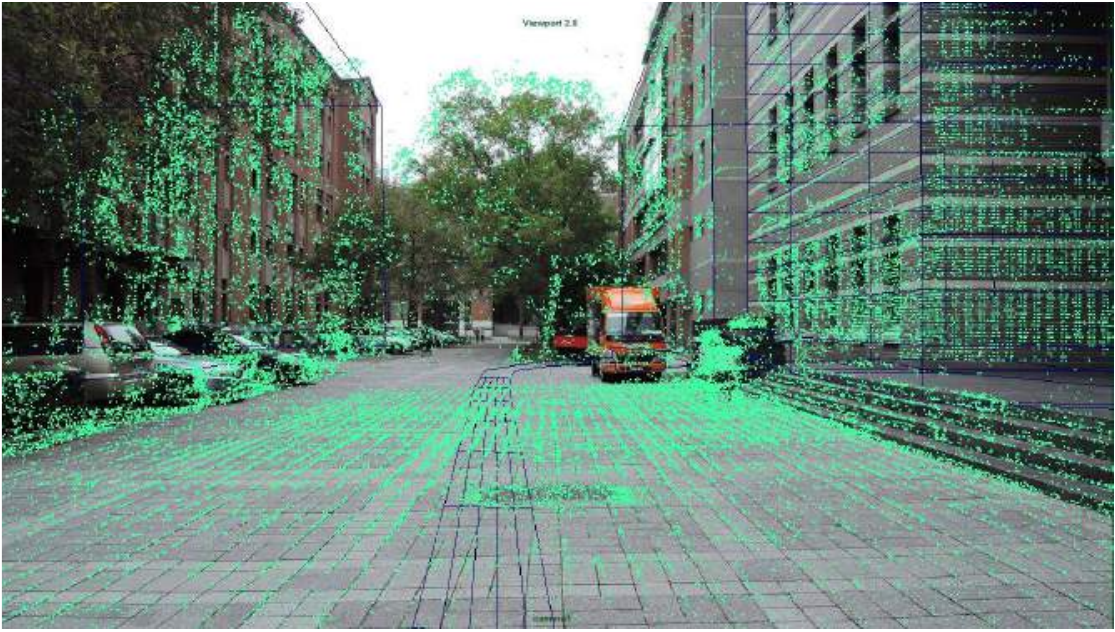}}
\subfigure[]{
\includegraphics[width=0.7\linewidth]{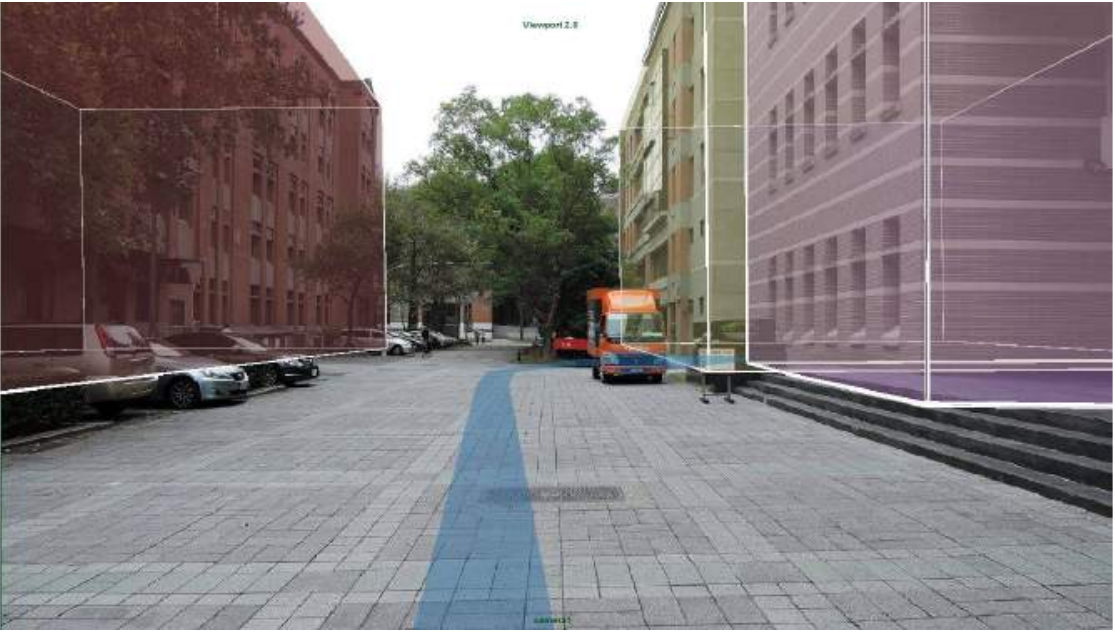}}
\end{center}
   \caption{Results of positioning for AR display: (a) point cloud and structures of buildings on Maya, (b) image with projected point cloud, and (c) a result of AR display.}
\label{fig:ARDisplay}
\end{figure}

To demonstrate the proposed positioning method can help developing AR display, we load the point cloud model of Scene \#4 to Maya and draw the corresponding structures of buildings and AR information on it, as shown in Figure~\ref{fig:ARDisplay}(a).  We then load the camera poses of each image estimated by our method to Maya to render the projection image. Figure~\ref{fig:ARDisplay}(b) is the result of projecting the point cloud onto an image, and Figure~\ref{fig:ARDisplay}(c) show the result of AR display. As we can see, the AR information can be aligned to the image well and our method providing highly accurate estimation is helpful for future AR applications.

\section{Conclusions}

We have proposed a novel algorithm for the IoT framework to make vision-based positioning practical in real situations.  
We formulate model compression as a weighted k-cover problem to condense the model while preserving the scene structure. 
Experimental results show that sub-meter positioning accuracy can be achieved based on the compressed models for both image and video tests. 
We also present a model update method for long-term data.
The released long-term positioning dataset will be helpful to the future progress of vision-based positioning adaptive to scene changes.

\section{Acknowledgments}
This work was supported by Ministry of Science and Technology, National Taiwan University and Intel Corporation under Grants MOST 103-2911-I-002-001 and NTU-ICRP-104R7501.

%
\bibliographystyle{abbrv}
\bibliography{PositioningReference}  

\begin{thebibliography}{10}

\bibitem{Arth}
C.~Arth, D.~Wagner, M.~Klopschitz, A.~Irschara, and D.~Schmalstieg.
\newblock Wide area localization on mobile phones.
\newblock {\em ISMAR}, 2009.

\bibitem{Brubaker}
M.~Brubaker, A.~Geiger, and R.~Urtasun.
\newblock Lost! leveraging the crowd for probabilistic visual
  self-localization.
\newblock {\em CVPR}, 2013.

\bibitem{Cao14}
S.~Cao and N.~Snavely.
\newblock Minimal scene descriptions from structure from motion models.
\newblock {\em CVPR}, 2014.

\bibitem{Chen}
D.~M. Chen, G.~Baatz, K.~Koser, S.~S. Tsai, R.~Vedantham, T.~Pylvanainen,
  K.~Roimela, X.~Chen, J.~Bach, M.~Pollefeys, B.~Girod, and R.~Grzeszczuk.
\newblock City-scale landmark identification on mobile devices.
\newblock {\em CVPR}, 2011.

\bibitem{Debevec}
P.~Debevec, C.~Taylor, and J.~Malik.
\newblock Modeling and rendering architecture from photographs: a hybrid
  geometry- and image-based approach.
\newblock {\em ACM SIGGRAPH}, 1996.

\bibitem{Driver}
T.~Driver.
\newblock Long-term prediction of gps accuracy: Understanding the fundamentals.
\newblock {\em ION GNSS International Technical Meeting of the Satellite
  Division}, 2007.

\bibitem{GoogleTango}
Google.
\newblock Google project tango.
\newblock \url{https://www.google.com/atap/projecttango}.

\bibitem{Gu}
Y.~Gu, A.~Lo, and I.~Niemegeers.
\newblock A survey of indoor positioning systems for wireless personal
  networks.
\newblock {\em IEEE COMST}, 11(1):13--32, 2009.

\bibitem{Hartley}
R.~Hartley and A.~Zisserman.
\newblock Multiple view geometry in computer vision.
\newblock {\em Cambridge University Press}, 2004.

\bibitem{Irschara}
A.~Irschara, C.~Zach, J.~Frahm, and H.~Bischof.
\newblock From structure-from-motion point clouds to fast location recognition.
\newblock {\em CVPR}, 2009.

\bibitem{Johns13}
E.~Johns and G.~Z. Yang.
\newblock Dynamic scene models for incremental, long-term, appearance-based
  localisation.
\newblock {\em ICRA}, 2013.

\bibitem{Johns13Feature}
E.~Johns and G.~Z. Yang.
\newblock Feature co-occurrence maps: Appearance-based localisation throughout
  the day feature co-occurrence maps: Appearance-based localisation throughout
  the day.
\newblock {\em ICRA}, 2013.

\bibitem{Kalman}
R.~Kalman.
\newblock A new approach to linear filtering and prediction problems.
\newblock {\em Journal of Basic Engineering}, 82(1):35--45, 1960.

\bibitem{Kneip}
L.~Kneip, M.~Chli, R.~Siegwart, R.~Siegwart, and R.~Siegwart.
\newblock Robust real-time visual odometry with a single camera and an imu.
\newblock {\em BMVC}, 2011.

\bibitem{Koyuncu}
H.~Koyuncu and S.~H. Yang.
\newblock A survey of indoor positioning and object locating systems.
\newblock {\em IJCSNS}, 10(5):121--128, 2010.

\bibitem{Lategahn}
H.~Lategahn and C.~Stiller.
\newblock Vision-only localization.
\newblock {\em IEEE T-ITS}, 15(3):1246--1257, 2014.

\bibitem{Li10}
Y.~Li, N.~Snavely, and D.~Huttenlocher.
\newblock Location recognition using prioritized feature matching.
\newblock {\em ECCV}, 2010.

\bibitem{Li12}
Y.~Li, N.~Snavely, D.~Huttenlocher, and P.~Fua.
\newblock Worldwide pose estimation using 3d point clouds.
\newblock {\em ECCV}, 2012.

\bibitem{Lim}
H.~Lim, S.~Sinha, M.~Cohen, and M.~Uyttendaele.
\newblock Real-time image-based 6-dof localization in large-scale environments.
\newblock {\em ISMAR}, 2012.

\bibitem{Liu07}
H.~Liu, H.~Darabi, P.~Banerjee, and J.~Liu.
\newblock Survey of wireless indoor positioning techniques and systems.
\newblock {\em IEEE T-SMC-C}, 37(6):1067--1080, 2007.

\bibitem{Liu12}
H.~Liu, T.~Mei, J.~Luo, H.~Li, and S.~Li.
\newblock Finding perfect rendezvous on the go: Accurate mobile visual
  localization and its applications to routing.
\newblock {\em ACM Multimedia}, 2012.

\bibitem{Lowe}
D.~Lowe.
\newblock Distinctive image features from scale-invariant keypoints.
\newblock {\em IJCV}, 60(2):91--110, 2004.

\bibitem{Martin10}
E.~Martin, O.~Vinyals, G.~Friedland, and R.~Bajcsy.
\newblock Precise indoor localization using smart phones.
\newblock {\em ACM Multimedia}, 2010.

\bibitem{MSHoloLens}
Microsoft.
\newblock Microsoft hololens.
\newblock \url{http://www.microsoft.com/microsoft-hololens/en-us}.

\bibitem{Middelberg}
S.~Middelberg, T.~Sattler, O.~Untzelmann, and L.~Kobbelt.
\newblock Scalable 6-dof localization on mobile devices.
\newblock {\em ECCV}, 2014.

\bibitem{Modsching}
M.~Modsching, R.~Kramer, and K.~Hagen.
\newblock Field trial on gps accuracy in a medium size city: The influence of
  built-up.
\newblock {\em WPNC}, 2006.

\bibitem{Muja}
M.~Muja and D.~Lowe.
\newblock Fast approximate nearest neighbors with automatic algorithm
  configuration.
\newblock {\em VISAPP}, 2009.

\bibitem{Park}
H.~S. Park, Y.~Wang, E.~Nurvitadhi, J.~C. Hoe, Y.~Sheikh, and M.~Chen.
\newblock 3d point cloud reduction using mixed-integer quadratic programming.
\newblock {\em CVPR Workshops}, 2013.

\bibitem{NavGateHUD}
Pioneer.
\newblock Navgate hud.
\newblock \url{http://www.pioneer.eu/eur/page/products/navgatehud.html}.

\bibitem{Sattler11}
T.~Sattler, B.~Leibe, and L.~Kobbelt.
\newblock Fast image-based localization using direct 2d-to-3d matching.
\newblock {\em ICCV}, 2011.

\bibitem{Sattler12}
T.~Sattler, B.~Leibe, and L.~Kobbelt.
\newblock Improving image-based localization by active correspondence search.
\newblock {\em ECCV}, 2012.

\bibitem{Schmid}
K.~Schmid and H.~Hirschmuller.
\newblock Stereo vision and imu based real-time ego-motion and depth image
  computation on a handheld device.
\newblock {\em ICRA}, 2013.

\bibitem{Snavely}
N.~Snavely, S.~Seitz, and R.~Szeliski.
\newblock Photo tourism: Exploring photo collections in 3d.
\newblock {\em ACM ToG}, 25(3):835--846, 2006.

\bibitem{Ventura}
J.~Ventura and T.~Hollerer.
\newblock Wide-area scene mapping for mobile visual tracking.
\newblock {\em ISMAR}, 2012.

\bibitem{Wendel}
A.~Wendel, A.~Irschara, and H.~Bischof.
\newblock Natural landmark-based monocular localization for mavs.
\newblock {\em ICRA}, 2011.

\bibitem{Yang}
M.~Y. Yang and W.~Forstner.
\newblock Plane detection in point cloud data.
\newblock {\em MCG}, 2010.

\bibitem{Zamir}
A.~Zamir and M.~Shah.
\newblock Accurate image localization based on google maps street view.
\newblock {\em ECCV}, 2010.

\end{thebibliography}

\end{document}